\documentclass{article}
\PassOptionsToPackage{numbers, compress}{natbib}
 \usepackage[main, final]{neurips_2026}
\usepackage[utf8]{inputenc} % allow utf-8 input
\usepackage[T1]{fontenc}    % use 8-bit T1 fonts
\usepackage{hyperref}       % hyperlinks
\usepackage{url}            % simple URL typesetting
\usepackage{booktabs}       % professional-quality tables
\usepackage{amsfonts}       % blackboard math symbols
\usepackage{nicefrac}       % compact symbols for 1/2, etc.
\usepackage{microtype}      % microtypography
\usepackage{xcolor}         % colors
\usepackage{fontawesome5}
\usepackage{graphicx}
\usepackage{amsmath}
\usepackage{multirow}
\usepackage{multicol}
\usepackage{algorithm}
\usepackage{algorithmic}
\usepackage[table]{xcolor}
\usepackage{pifont}   % 用于 \ding 符号
\usepackage{xcolor}   % 用于颜色
\usepackage{colortbl} % 用于表格背景色
\usepackage{wrapfig}
\usepackage{enumitem}
\newcommand{\cmark}{\textcolor{green}{\ding{51}}} % 定义绿色的勾
\newcommand{\xmark}{\textcolor{red}{\ding{55}}}   % 定义红色的叉
\definecolor{deeppurple}{RGB}{98,0,238}

\title{From Clips to Streams: A Unified Framework for Streaming Sign Language Translation}

\usepackage{color}

\author{%
Hoang Quan Dang$^{1}$  \quad Xin Shen$^{1,2}$ \quad Yanbin Liu$^{3}$ \quad \textbf{Xin Yu}$^{2}\thanks{Corresponding author.}~$ \quad Ling Chen$^{3}$\\
\\
$^{1}$~School of Electrical Engineering and Computer Science, The University of Queensland \\
$^{2}$~Australian Institute for Machine Learning, Adelaide University \\
$^{3}$~Australian Artificial Intelligence Institute, University of Technology Sydney \\
\texttt{q.dang@uq.edu.au}
}

\begin{document}

\maketitle

\begin{abstract}
Sign Language Translation (SLT) research has typically focused on pre-segmented clips, a constraint that disconnects models from the continuous reality of real-world communication.
To bridge the gap from clips to streams, we formally define \textbf{Streaming SLT}, a realistic task demanding the joint discovery and translation of linguistic events from continuous, untrimmed video inputs.
To address this challenge, we propose \textbf{StreamSLST}, the first end-to-end framework for \textbf{Stream}ing \textbf{S}ign \textbf{L}anguage \textbf{S}egmentation and \textbf{T}ranslation via Deformable Transformer architecture. 
Unlike cascaded approaches that suffer from error propagation, \textbf{StreamSLST} employs a parallel decoding strategy to jointly optimize \textit{temporal segmentation} and \textit{translation} in a single pass. 
To enable scalable training on long-form videos, we incorporate a \textit{Sentence-aware Sliding Window Sampler} that crops continuous streams into trainable units. 
We further design a \textit{Tri-modal Visual-Language Pretraining} stage that aligns pose dynamics with textual semantics before joint optimization.
To support comprehensive evaluation, we conduct experiments on two real-world streaming datasets, BOBSL and How2Sign, together with two newly constructed synthetic streaming benchmarks, \textit{Streaming-Phoenix-2014T} and \textit{Streaming-CSL-Daily}. The results demonstrate that \textbf{StreamSLST} consistently outperforms a matched cascaded segmentation-then-translation pipeline. 
Interestingly, our analysis reveals that joint optimization learns wider, translation-effective temporal support that simple expansion or rescaling of the annotated spans fails to reproduce. 
Our work establishes the first robust benchmark for continuous sign language understanding and paves the way for accessible real-world communication interfaces. The source code and data are available at ~\href{https://github.com/streamsl/StreamSLST}{\faGithub~\textcolor{deeppurple}{StreamSLST}}.

\end{abstract}

\begin{figure*}[h] % Positioning: h = here
    \centering
    \includegraphics[width=0.99\textwidth]{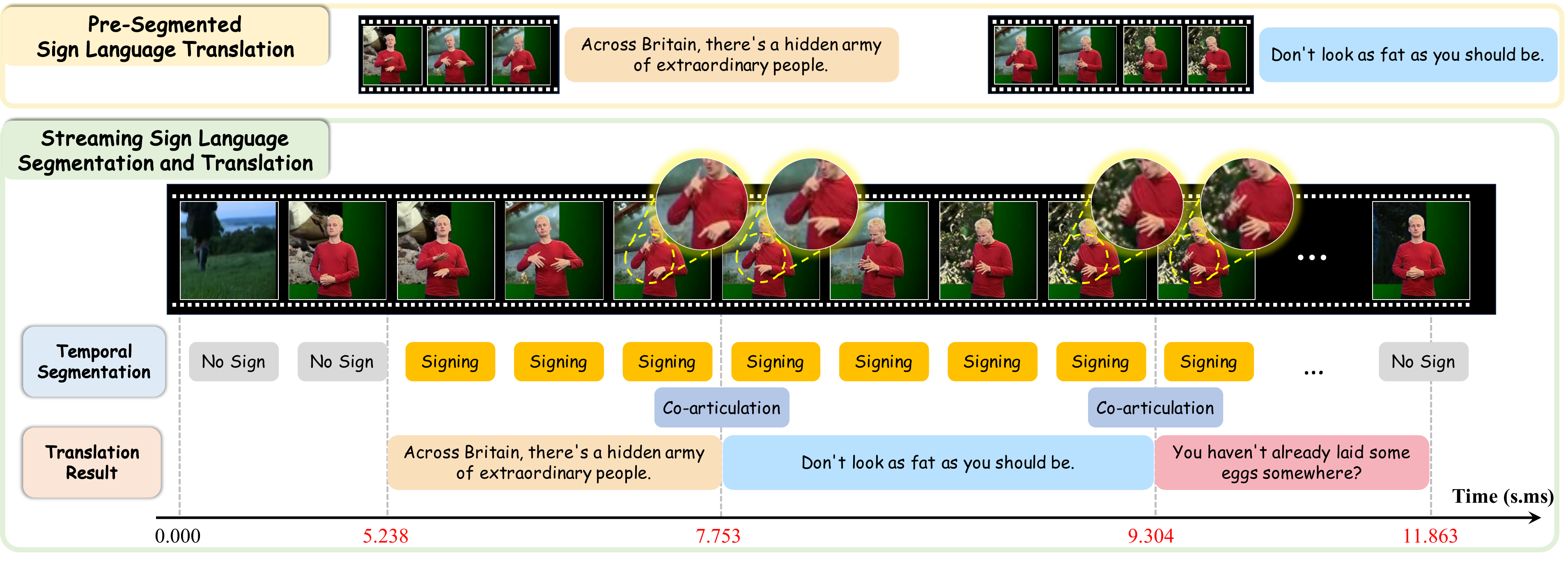}
    \caption{Comparison between Pre-Segmented SLT and the proposed \textbf{Streaming SLT} task. Unlike traditional clip-based methods (top), \textbf{Streaming SLT} (bottom) processes continuous streams, requiring the model to autonomously handle two critical challenges: 
    (1) Activity Detection, distinguishing signing intervals from non-signing transitions (e.g., the signer at rest, scene changes, or speaker close-ups); and 
    (2) Boundary Ambiguity, where co-articulation visually blurs the transition between adjacent signing sentences.
    }
    \label{fig:slt}
\end{figure*}

\section{Introduction}
\label{sec:intro}
Effective communication is the foundation of human interaction, yet for more than 1.5 billion people living with some degree of hearing loss today~\citep{who,ke2026ivqald}, digital barriers remain pervasive. 
Sign language serves as a crucial linguistic bridge, employing a sophisticated grammar of manual and non-manual signals~\citep{sandler2006,pfau2010}.
While platforms such as Microsoft Teams provide accessibility features for sign language users, they do not provide automatic sign-language-to-text translation, and access therefore still depends on human interpreters~\citep{moryossef2020real,shen2025auslanweb,sheng2026mobile,sheng2024ai}.
The resulting dependency on human interpreters further creates a bottleneck for accessibility, limiting the Deaf and Hard-of-Hearing community's autonomy in professional and social spheres~\citep{shen2024mm}.

While the field of Sign Language Translation (SLT) has witnessed rapid advancements~\citep{camgoz2018neural, unisign, gfslt,shen2023auslan,shen2026banz}, existing solutions are ill-equipped for continuous, untrimmed signing flow of real-world communications.
The majority of state-of-the-art models rely on the artificial constraint of processing pre-segmented clips where the start and end of each sentence are known \textit{in advance}.
This reliance makes them impractical for streaming applications, where the input is an untrimmed video stream containing indeterminate sentence boundaries and non-signing intervals. 
To bridge the gap from clips to streams, we formally define \textit{\textbf{Streaming} \textbf{S}ign \textbf{L}anguage \textbf{T}ranslation} (\textbf{Streaming SLT}), a unified task to jointly localize linguistic events and translate their content from continuous streams.
As illustrated in Figure~\ref{fig:slt}, unlike traditional SLT, this realistic setting imposes two critical challenges: (i) \textit{Activity Detection}, which requires precisely distinguishing valid signing segments from background noise; and (ii) \textit{Boundary Ambiguity}, arising from the rapid and diverse co-articulation~\citep{zuo2024towards} that visually blurs the transitions between sentences.

To address these challenges, we propose the first end-to-end framework, \textbf{StreamSLST}, designed explicitly for \textbf{Streaming SLT} task via joint \textit{segmentation} and \textit{translation}.
Instead of relying on cascaded modules, \textbf{StreamSLST} presents a unified architecture tailored to the skeleton modality that processes continuous 2D pose streams directly. 
Our model leverages group-specific normalized pose features \citep{cosign} and spatial-temporal graph convolutions (ST-GCN) \citep{stgcn,shen2025cross} to capture fine-grained signing dynamics. 
A Deformable Transformer with parallel decoding heads \citep{pdvc} then jointly predicts temporal boundaries and translated sentences in a single pass. % 
To enable scalable training on massive unsegmented datasets, we further introduce a \textit{Sentence-aware Sliding Window Sampler} that crops continuous streams into trainable units, each containing at least one valid sentence. 
To cope with the resulting noisy window-level supervision, we split training into two stages: a \textit{Tri-modal Visual-Language Pretraining} stage that grounds the pose backbone in linguistic semantics using paired text, followed by an end-to-end stage that jointly optimizes localization and translation. 

We empirically validate \textbf{StreamSLST} across diverse streaming scenarios. In addition to two real-world streaming datasets, BOBSL and How2Sign, we construct two synthetic streaming benchmarks, \textit{Streaming-CSL-Daily} and \textit{Streaming-Phoenix-2014T}, by converting pre-segmented SLT datasets into continuous streams. 
Experiments on these four benchmarks show that \textbf{StreamSLST} consistently outperforms a matched cascaded segmentation-then-translation pipeline on identical predicted intervals.
Our main contributions are summarized as follows: 
\begin{itemize}[leftmargin=15pt]
\item We define the \textbf{Streaming SLT} task and propose \textbf{StreamSLST}, a unified end-to-end framework that jointly optimizes temporal segmentation and translation from continuous streams via a Deformable Transformer. 
\item We enable scalable training on long-form videos by introducing a \textit{Sentence-Aware Sliding Window Sampler}, complemented by \textit{Tri-modal Visual-Language Pretraining} to ensure semantic alignment. 
\item We provide a comprehensive evaluation protocol for \textbf{Streaming SLT}, covering two real-world long-form datasets and two newly constructed synthetic streaming benchmarks, \textit{Streaming-Phoenix-2014T} and \textit{Streaming-CSL-Daily}.
\item We establish the first benchmark on four \textbf{Streaming SLT} datasets and analyze how joint training and temporal-support choices affect localization and translation. 
\end{itemize}

\section{Related Work}
\textbf{Sign Language Translation (SLT)}. 
SLT generates spoken language directly from sign videos. 
Early gloss-based systems first recognized a gloss sequence and then translated it~\citep{phoenix}. \citet{camgoz2020sign} removed that two-step pipeline by training recognition and translation together under a CTC objective, while gloss supervision remained a requirement.
Gloss supervision faces two fundamental limitations: (1) heavy reliance on scarce gloss annotations, which require sign language expertise \citep{li2020tspnet} and (2) an information bottleneck where discrete glosses discard critical non-manual cues \citep{gfslt}. 
% Adapting such modular systems to streaming environments necessitates disjoint multi-stage processing, inherently introducing latency and severe error propagation.
To mitigate these limitations, gloss-free SLT utilizes end-to-end architectures mapping video directly to text. 
Recent advances leverage massive pre-training \citep{unisign,shen2026diverse,scalingslt} or bridge visual encoders with frozen LLMs \citep{mmslt,shen2026diverse,shen2026fingercap,sign2gpt} to enhance semantic understanding. 
Another pivotal trend is Visual-Language Pretraining (VLP), where GFSLT-VLP \citep{gfslt} demonstrates that aligning visual features with textual embeddings via contrastive objectives effectively compensates for the lack of gloss supervision. 

To better integrate LLMs and bridge the visual-textual gap, existing works explore factorized two-stage learning~\cite{chen2024factorized}, pseudo-gloss pre-training with lightweight adapters~\cite{sign2gpt}, and the transformation of continuous sign videos into discrete language-like tokens~\cite{gong2024llms}. 
To enhance visual representation learning and cross-modal alignment, other methods also propose visual alignment pre-training~\cite{jiao2024visual}, or joint content and context representation learning~\cite{chen2025c}.
% and unified generative pre-training across various downstream tasks~\cite{unisign}. 
Despite these architectural strides, a critical limitation persists across both gloss-based and gloss-free SLT: they operate exclusively on pre-segmented clips with oracle sentence boundaries.  Each system reads one clip whose start and end somebody has already chosen, so this literature answers what a sentence says once its extent is known and leaves open how that extent is found. For instance, Scaling SLT~\citep{scalingslt} uses caption alignment as a pretraining task and its downstream evaluation assumes pre-segmented sentence clips and does not evaluate sentence discovery, non-signing interval rejection, or temporal localization. This dependency renders them ill-suited for \textbf{Streaming SLT}, which demands the autonomous disentanglement of linguistic events from continuous, untrimmed streams.

\textbf{Temporal Segmentation in Sign Language}. 
Continuous signing lacks clear pauses, complicating segmentation into signs, phrases, or sentences. Therefore, temporal segmentation, which identifies the boundaries of meaningful linguistic units within continuous signing, is a prerequisite for \textbf{Streaming SLT} but has received limited attention as a standalone task. Early work on this matter employed sequence labeling approaches using BIO tagging (Begin-Inside-Outside) on optical flow, demonstrating superior performance for continuous sequences \citep{bioseg}. Another recent study applies temporal segmentation to BOBSL \citep{bobsl}, using an attention-based sequence-to-sequence model to predict subtitle-level boundaries \citep{rao-etal-2025-sign}. They trained on a small, manually-aligned subset of 40 source videos with 375-frame (15-second) windows and evaluated segmentation quality via average IoU and alignment accuracy. 
While this work represents a closely related prior art, it has two critical limitations: (1) the segmentation tasks are not coupled with the translation, requiring separate models and sequential processing in production environments, and (2) the dataset size is extremely limited, with only 40 source videos, which restricts the model's capacity to generalize.
Other related segmentation work also only returns boundaries without producing text~\citep{bull2020,renz2021}. On the other hand, alignment methods receive both video and target text and return timings for that text \citep{bull2021,sea2026}. 
Unlike the above work, our approach unifies segmentation and translation in a single end-to-end framework.

\textbf{Incremental SLT}.
This line of work shortens delays within one utterance. SimulSLT~\citep{simulslt} decodes a pre-segmented utterance with a wait-$k$ policy and gloss-supervised boundaries, and Online SLT~\citep{zuo2024towards} slides a sign-level recognizer over continuous signing and translates the recognized glosses. Neither discovers a variable number of sentence events from untrimmed input. 
In \textbf{Streaming SLT}, \textit{streaming} describes the input rather than the output delay: the task receives untrimmed video alone and asks for the extent and the text of every sentence together, without gloss annotation (Appendix~\ref{app:positioning}).
Our task specifies what a system must output and leaves its latency open, and \textbf{StreamSLST} emits a sentence once the 15-second input window that contains it has arrived (Appendix~\ref{app:cost}). 

\section{Task Formulation}
\label{sec:task}

In the \textbf{Streaming SLT} task, we treat the input not as pre-segmented clips but as a continuous, untrimmed video stream. 
The objective is to jointly localize the temporal boundaries of signing events and translate their linguistic content within this continuous flow.

Formally, given a long-form sign language video $\mathbf{V} = \{v_1, v_2, \ldots, v_T\}$ with $T$ frames, the goal is to predict a set of $M$ temporally-grounded linguistic events, where $M$ is unknown in advance:
\begin{equation}
\mathcal{E} = \{(s_i, e_i, \mathbf{c}_i) \mid i = 1, 2, \ldots, M\},
\end{equation}
where $s_i, e_i \in [1, T]$ represent the start and end frame indices of the $i$-th sentence\footnote{``Sentence'' refers to one subtitle-aligned linguistic unit (\emph{e.g.}, one English caption in BOBSL/How2Sign).}, satisfying $s_i < e_i$. The term $\mathbf{c}_i$ denotes the corresponding spoken language translation (\emph{e.g.}, an English sentence), which is a sequence of words derived from a target vocabulary $\mathcal{V}$. 
Each event corresponds to a semantically complete linguistic unit, effectively filtering out non-signing transition states. 

\section{Methodology}

\subsection{End-to-End StreamSLST}
\label{sec:architecture}

\begin{figure*}[t]
    \centering
    \includegraphics[width=1\textwidth]{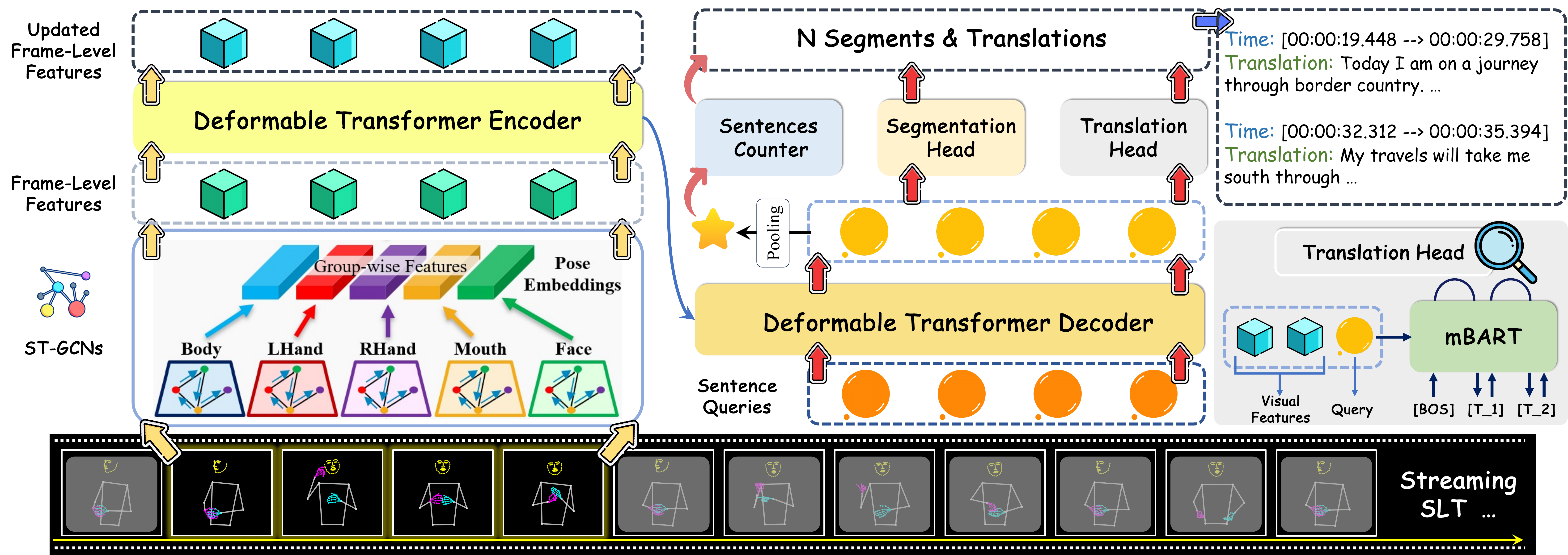}
    \caption{Overview of our \textbf{StreamSLST} framework. 
    A Deformable Transformer encodes pose features and refines learnable Sentence Queries via deformable cross-attention. 
    The refined queries drive three parallel heads: a Segmentation Head, an auxiliary Sentences Counter, and a Translation Head.
    As magnified on the right, the Translation Head fuses the specific Sentence Query with sampled Visual Features to condition the mBART decoder.
    }
    \label{fig:pdvc}
\end{figure*}

We propose \textbf{StreamSLST}, a unified end-to-end framework that jointly segments and translates sign language sentences from continuous pose streams. 
As illustrated in Figure \ref{fig:pdvc}, the architecture comprises three core components: (1) a spatial-temporal pose embedding backbone, (2) a Deformable Transformer Encoder-Decoder, and (3) parallel prediction heads for joint segmentation and translation.

\textbf{Spatial-Temporal Pose Embedding.}
The foundation of our visual encoder is an ST-GCN \citep{stgcn} that transforms normalized pose sequences into semantically rich feature representations. 
Given an input window of $T$ frames with 77 selected keypoints organized into five anatomical groups (body, left+right hands, face, mouth), the pose sequence is represented as $\mathbf{P} \in \mathbb{R}^{B \times T \times 77 \times 3}$.
Following \cite{cosign}, we construct separate graph topologies for each anatomical group. Spatial convolutions aggregate features from neighboring keypoints, while temporal convolutions capture motion dynamics. After processing through ST-GCN blocks, we obtain 5 group-specific feature sequences. These are spatially aggregated and concatenated to yield $\mathbf{F}_\text{concat}$.
A final linear projection maps this representation to the Transformer's hidden dimension, producing unified pose features \(\mathbf{F}_{\text{pose}}\).

\textbf{Deformable Transformer Encoder--Decoder.}
Building upon the pose features, we adopt the Deformable Transformer Encoder--Decoder of \cite{deformdetr,pdvc} as our temporal localization backbone. This choice is deliberate: existing fully-attentive Transformers scale quadratically with sequence length, which is prohibitive for the long pose windows required by streaming SLT, while Deformable Attention reduces the complexity to a sparse set of sampling points around learned reference locations. To the best of our knowledge, this is the first work that applies a 1D temporal-deformable formulation to sign-language streaming and extends it with a translation head conditioned on per-query reference points.
Given multi-scale temporal feature maps \(\mathbf{X} = \{\mathbf{x}^l\}_{l=1}^L\) where \(\mathbf{x}^l \in \mathbb{R}^{C \times T_l}\), a query element \(\mathbf{q}_j\), and a normalized temporal reference point \(p_j \in [0, 1]\), the Multi-Scale Deformable Attention (MSDAtt) outputs a context vector via:
\begin{equation}
\label{eq:deform_attn}
\text{MSDAtt}(\mathbf{q}_j, p_j, \mathbf{X}) = \sum_{l=1}^L \sum_{k=1}^K A_{jlk} \mathbf{W} \mathbf{x}^l(p_j + \Delta p_{jlk}),
\end{equation}
where $l$ indexes the feature level, and $k$ indexes the sampling point. 
\(\mathbf{W}\) is the projection matrix. \(A_{jlk} \in [0, 1]\) denotes the attention weight, and \(\Delta p_{jlk} \in \mathbb{R}\) represents the learned sampling offset relative to the reference point \(p_j\). 
Both \(A_{jlk}\) and \(\Delta p_{jlk}\) are obtained via linear projections of the query \(\mathbf{q}_j\). Bilinear interpolation is used to sample features at the fractional location \(p_j + \Delta p_{jlk}\).

The \textit{Encoder} stacks layers of MSDAtt to enrich the pose features into context-aware temporal representations. The \textit{Decoder} introduces \(N\) learnable \textit{Sentence Queries} \(\mathbf{Q} = \{\mathbf{q}_i\}_{i=1}^N\). 
Each query serves as an anchor to probe the encoder outputs via MSDAtt (Cross-Attention), refining its representation to predict sentence boundaries and translation content.

\textbf{Parallel Prediction Heads.}
From decoder outputs, multiple parallel heads decode complementary aspects of each sentence query.

\textit{Sentences Counter.} 
To explicitly model the sentence density 
within the window, we introduce an auxiliary Sentences Counter. 
Unlike individual query predictions, this head operates on the global context. 
We first apply max-pooling over all learnable sentence queries to obtain a unified representation. 
This pooled feature is then fed into a classification head to predict the total number of valid linguistic events $M$ in the current video stream. 

\textit{Segmentation Head.} 
For each sentence query $\mathbf{q}_i$, a three-layer MLP predicts the temporal boundaries \(\hat{\mathbf{b}}_i = (\hat{c}_i, \hat{w}_i) \in [0, 1]^2\), and a linear classifier outputs the foreground logit \(\hat{y}_i\) that gates whether the query corresponds to a real sentence. 
The \textit{Segmentation Head} is independent of the \textit{Sentence Counter} (not stacked). They receive complementary supervision: the per-query foreground classifier is supervised with focal loss, the boundary predictor with GIoU on the matched query--sentence pair, and the \textit{Sentence Counter} with the BCE loss on the one-hot sentence count defined below.

\textit{Context-Aware Translation Head.} 
This head is where our design departs from dense video captioning (DVC). Previous work~\citep{pdvc} trained an LSTM captioner from scratch with re-sampled encoder features for every generated word through deformable soft attention. We translate with a pretrained multilingual decoder instead, mBART~\citep{mbart}, which brings its own language prior to our target languages. We thus give mBART localized visual evidence once per query as a single fused state and rely on its prior to generate the sentence. 
Recall that each \textit{Sentence Query} $\mathbf{q}_i$ is one of the $N$ learnable embeddings refined by the decoder, it carries an aggregated, query-centric summary of the input but does \emph{not} retain frame-level visual detail, which is essential for translating mouthings, fingerspelling, and other fine-grained cues. We therefore condition translation on \emph{both} the query and a localized \textit{Visual Features} drawn from the Encoder. 
For each query $q_i$, we read its predicted center $\hat{c}_i$, convert it to integer frame indices at each pyramid level, and \emph{gather} (nearest-neighbor) the corresponding feature vectors. Nearest-neighbor sampling is used in place of bilinear interpolation because $\hat{c}_i$ is already a learned, refined location -- adding a second smoothing step would blur the query's own localization decision. 
Features from all levels are averaged into $\mathbf{v}_i$ to capture both fine and coarse temporal context, and the conditioned hidden state is the linear fusion $\mathbf{h}_i = \mathbf{W}_{\text{proj}}[\mathbf{v}_i; \mathbf{q}_i]$.
This context-aware state \(\mathbf{h}_i\) is then fed into the mBART \citep{mbart} decoder to autoregressively generate the target sentence tokens \([\text{BOS}, T_1, T_2, \dots]\).

% \subsubsection{Joint Optimization Objective}
\textit{Joint Optimization Objective.}
Each of the $N$ sentence queries produces a candidate triple (foreground logit, boundary, translation), so the decoder emits a fixed set of $N$ candidates per window. The ground truth, however, contains a variable number $M \leq N$ of sentences. To supervise this set-to-set mapping, we use \textbf{Hungarian Matching}~\citep{hungarian}, which assigns each GT sentence to the unique candidate that minimizes the pairwise cost below. Unmatched candidates are trained as background.
Based on the optimal matching, the total loss \(\mathcal{L}_{\text{total}}\) combines classification, regression, and generation objectives:
\begin{equation}
    \mathcal{L}_{\text{total}} = \lambda_{\text{cls}} \mathcal{L}_{\text{focal}} + \lambda_{\text{giou}} \mathcal{L}_{\text{giou}} + \lambda_{\text{count}} \mathcal{L}_{\text{bce}} + \lambda_{\text{trans}} \mathcal{L}_{\text{text}}.
    \label{eq:joint_loss}
\end{equation}
We detail the components as follows: 
(1) for \textit{foreground classification}, a \textbf{Sigmoid Focal} loss (\(\mathcal{L}_{\text{focal}}\)) is applied to address the severe class imbalance between a few positive signing queries and many background queries; 
(2) the \textit{localization} loss (\( \mathcal{L}_{\text{giou}}\)) is \textbf{Generalized IoU (GIoU)} loss to impose scale-invariant constraints on the predicted temporal segments; 
(3) the \textit{counter head} is supervised via a \textbf{Binary Cross Entropy} loss ($\mathcal{L}_{\text{bce}}$) over a one-hot encoding of the ground-truth sentence count $M \in \{0, 1, \ldots, N_{\max}\}$, where $N_{\max}$ is the maximum allowed sentences per window. This auxiliary objective complements the per-query classification: the $N$ queries individually decide whether they correspond to a sentence, while the counter directly regularizes how many of them should fire, preventing the over- or under-prediction failure modes common to detection transformers~\citep{deformdetr};
(4) for the \textit{translation head}, we minimize the \textbf{Negative Log-likelihood} (\(\mathcal{L}_{\text{text}}\)) over the matched target sequence \(\mathbf{c}_i = (c_1^i, \ldots, c_{L_i}^i)\). This translation loss is formulated as:
\begin{equation}
    \mathcal{L}_{\text{text}}^i = -\sum_{t=1}^{L_i} \log P(c_t^i \mid c_{<t}^i, \mathbf{h}_i),
\end{equation}
where $P(\cdot)$ denotes the probability distribution over the vocabulary produced by the mBART decoder.

\subsection{Training Strategies}
\label{sec:training_strategy}

Training \textbf{StreamSLST} on real-world datasets presents two fundamental challenges: (1) the difficulty of processing long-form videos, and (2) the lack of gloss-level supervision. 
To address these, we introduce a \textit{Sentence-Aware Sliding Window Sampler} and a \textit{Two-Stage Optimization Paradigm}.

% \subsubsection{Sentence-Aware Sliding Window Sampler}
\textbf{Sentence-Aware Sliding Window Sampler.}
To handle the data scarcity and long video streams, we employ a dynamic sampling approach (Appendix Algorithm~\ref{alg:sliding_window}). This sampler operates differently in training and evaluation -- both are further illustrated in Appendix Figure ~\ref{fig:sampling}.

% \textit{Training: On-the-fly Stochastic Sampling.} 
% We randomly sample windows of size \(W\). This stochasticity can act as a form data augmentation, exposing the model to diverse temporal contexts. To ensure efficiency, we impose a constraint: each window must contain at least one valid sentence\XY{a gloss?}. If random sampling fails, we employ a fallback strategy that greedily selects the ``densest'' window using a two-pointer sweep algorithm.

% \textit{Evaluation: Deterministic Fixed-Stride.} 
% For validation, we adopt a deterministic approach with a fixed stride to ensure complete coverage of the video, preventing ``blind spots'' between windows.\XY{this paragraph is not clear. why do you suddenly mention evaluation? I guess you need to show some figures to explain this idea.}

\textit{Training: stochastic, on-the-fly.} For each video, we draw a random window of size $W$ and require it to contain at least one fully-enclosed sentence whose duration lies in the valid range $[1\,\text{s}, 20\,\text{s}]$~\citep{litfic}. If the random draw fails after a small number of attempts, we fall back to a two-pointer sweep that returns the densest valid window. Because each video yields a different window every epoch, the sampler doubles as a strong augmentation: the model sees the same content under many temporal crops, and never overfits to a single fixed segmentation of the corpus.

\textit{Evaluation: deterministic fixed-stride.} Stochastic sampling is incompatible with reproducible evaluation; we therefore precompute, once per dataset, an exhaustive list of overlapping fixed-stride windows with stride equal to a fixed fraction of $W$. The overlap is to avoid ``blind spots'' in which a sentence falls between two adjacent non-overlapping windows and is never seen by the model.

\textbf{Two-Stage Optimization Paradigm.} 
To align pose dynamics with textual semantics, we propose a Two-Stage Optimization Paradigm.

\textit{Stage 1: Tri-modal Visual-Language Pretraining.} 
Motivated by \cite{gfslt} but tailored to pose-only inputs, this stage aligns three views of the same window through contrastive learning. We freeze the Deformable Transformer and train only the ST-GCN backbone (and a small text projector) on the following three modality vectors per window:
\begin{itemize}[leftmargin=*]
\item $\mathbf{v}_1, \mathbf{v}_2 \in \mathbb{R}^d$: two \emph{masked pose views} of the same window, obtained by randomly partitioning anatomical-group sub-windows into two disjoint sets and zeroing one set per view (following the complementary masking of~\cite{cosign}). The two views thus carry complementary visual information from the same input, providing a view-level positive pair without requiring different videos.
\item $\mathbf{t} \in \mathbb{R}^d$: a \emph{window-level text vector} produced by tokenizing the concatenation of all sentence translations inside the window, embedding the tokens through a learnable layer, and mean-pooling them through a small projection head.
\end{itemize}
Given $B$ windows in a batch, all $3B$ vectors live in the same space; we treat the $(\mathbf{v}_1^i, \mathbf{v}_2^i)$, $(\mathbf{v}_1^i, \mathbf{t}^i)$, $(\mathbf{v}_2^i, \mathbf{t}^i)$ pairs from the same window $i$ as positives, and all cross-window pairs as in-batch negatives. The pairwise InfoNCE loss~\citep{infonce} with temperature $\tau$ is:
\begin{equation}
\mathcal{L}_{\text{InfoNCE}}(\mathbf{a}, \mathbf{p}) = -\log \frac{\exp(\text{sim}(\mathbf{a}^i, \mathbf{p}^i) / \tau)}{\sum_{j=1}^B \exp(\text{sim}(\mathbf{a}^i, \mathbf{p}^j) / \tau)}.
\end{equation}
% \XY{equation is a part of sentence. So it should be period.}
% Total contrastive loss aggregates 6 directional terms to enforce consistency across all modality pairs:
% \begin{equation}
%     \begin{aligned}
%     \mathcal{L}_{\text{contrastive}} = \frac{1}{6}[&\mathcal{L}_{\text{InfoNCE}}(\mathbf{v}_1, \mathbf{v}_2) + \mathcal{L}_{\text{InfoNCE}}(\mathbf{v}_2, \mathbf{v}_1) + \\
%     &\mathcal{L}_{\text{InfoNCE}}(\mathbf{v}_1, \mathbf{t}) + \mathcal{L}_{\text{InfoNCE}}(\mathbf{v}_2, \mathbf{t}) + \\
%     &\mathcal{L}_{\text{InfoNCE}}(\mathbf{t}, \mathbf{v}_1) + \mathcal{L}_{\text{InfoNCE}}(\mathbf{t}, \mathbf{v}_2)],
%     \end{aligned}
% \end{equation}
% \XY{I cannot even understand what they are...}

Using symmetric InfoNCE recipe of \citep{imagebind}, we instantiate the loss for both directions of each modality pair, giving 6 terms (3 pairs $\times$ 2 anchors). The total contrastive loss is their unweighted average:
\begin{equation}
    \mathcal{L}_{\text{contrastive}} = \frac{1}{6} \; \!\!\!\!\!\!\sum_{(\mathbf{a}, \mathbf{p}) \in \mathcal{P}}\!\!\!\!\!\mathcal{L}_{\text{InfoNCE}}(\mathbf{a}, \mathbf{p}),\;
    \mathcal{P} = \{(\mathbf{v}_1, \mathbf{v}_2), (\mathbf{v}_2, \mathbf{v}_1), (\mathbf{v}_1, \mathbf{t}), (\mathbf{t}, \mathbf{v}_1), (\mathbf{v}_2, \mathbf{t}), (\mathbf{t}, \mathbf{v}_2)\}.
\end{equation}
% We use $\tau = 0.1$ throughout, matching the ImageBind default.
% \textit{Stage 2: End-to-End Joint Training.} In the second stage, we unfreeze all parameters. This joint optimization enables the model to refine temporal proposals based on their semantic utility for translation, thereby creating synergy in which learned boundaries outperform GT annotations.
\textit{Stage 2: End-to-End Joint Training.} We simply unfreeze every parameter and resume training under the joint objective $\mathcal{L}_{\text{total}}$ from Equation~\ref{eq:joint_loss}. The role of $\mathcal{L}_{\text{contrastive}}$ ends with Stage 1 -- its purpose was to seed the ST-GCN backbone with pose--text alignment so that Stage 2 starts from a representation that already supports translation, rather than a randomly initialized one. Once joint training begins, the localization and translation heads can reshape the boundaries against the translation gradient itself, producing the synergy quantified in Section~\ref{sec:ablation}.

\section{Experiments}

\subsection{Dataset}
\label{sec:dataset}
We evaluate our approach on both real long-form sign language videos and controlled synthetic streaming data. For real-world evaluation, we use BOBSL~\cite{bobsl,rao-etal-2025-sign} and How2Sign~\cite{duarte2021how2sign}. 
In addition, we construct two synthetic streaming benchmarks, \textbf{Streaming-Phoenix-2014T} and \textbf{Streaming-CSL-Daily}, from the sentence-level Phoenix-2014T~\cite{camgoz2018neural} and CSL-Daily~\cite{csldaily} datasets.
These synthetic benchmarks provide exact oracle sentence boundaries, which are difficult to obtain reliably in real long-form videos.
Each synthetic stream is formed by concatenating multiple same-signer sentence clips with non-transcribed context and smooth transitions:
\begin{equation}
S = B_{\mathrm{pre}} \,\Vert\, c_1 \,\Vert\, \Delta_1 \,\Vert\, c_2 \,\Vert\, \cdots \, \Vert\, c_K \,\Vert\, B_{\mathrm{post}},
\end{equation}
where $c_i$ is a sentence-level sign clip, $\Delta_i$ is the transition between adjacent clips, and $B_{\mathrm{pre}}$, $B_{\mathrm{post}}$ are non-transcribed context segments.
The resulting datasets complement realistic but noisy long-form evaluation with controlled boundary supervision.
Further details are provided in Appendix~\ref{app:datasets}.

\subsection{Evaluation Metrics}

Following \cite{pdvc}, we computed temporal F1-score at multiple temporal IoU (tIoU) thresholds $\{0.3, 0.5, 0.7, 0.9\}$ to assess how well our \textbf{StreamSLST} framework localizes the signing segments across videos. A predicted sentence is considered a match if its temporal overlap with a GT sentence exceeds the threshold.
We then evaluated translation quality within those localized segments using five metrics: BLEU-4 \citep{bleu}, BLEURT \citep{bleurt}, ROUGE-L \citep{rouge}, CIDEr \citep{cider}, and METEOR \citep{meteor}.
Predictions are matched within each evaluation window, and Appendix~\ref{app:metrics} gives the full protocol.

\subsection{Experimental Setup}
\label{sec:exp_setup}
In Stage 1 contrastive pre-training, we disable all task-specific losses and optimize only \(\mathcal{L}_{\text{contrastive}}\). 
Following the best settings reported in \cite{imagebind} for image data, we train for 64 epochs with a learning rate of $5 \times 10^{-4}$ and a weight decay of 0.05. 
Then, in Stage 2 joint training, loss function weights follow the ratios established by \cite{pdvc}: $(\lambda_{\text{cls}}, \lambda_{\text{giou}}, \lambda_{\text{count}}, \lambda_{\text{trans}}) = (2, 4, 2, 2)$, where GIoU is weighted heavily to encourage accurate localization. 
We use a learning rate of $2 \times 10^{-4}$ and apply a weight decay of $1 \times 10^{-4}$ at this stage. 
All experiments are conducted on one NVIDIA A100 GPU with 80GB of memory, using AdamW optimizer \citep{adamw} with a linear decay scheduling over the training. 
We used 15-second pose windows, with 30 learnable sentence queries ($N=30$). 

\textbf{Baseline Configurations.}
To comprehensively validate our joint optimization approach, we establish two extra critical baselines. \emph{First}, we compare our translation architecture with the original parallel-decoding DVC formulation by \cite{pdvc} adopted in our framework. We use their recurrent decoding as a standard baseline for sentence-conditioned generation by replacing our mBART decoder with their deformable LSTM, keeping the full localization architecture and training recipe unchanged.
\emph{Second}, to assess whether \textbf{Streaming SLT} can be solved by a disjoint pipeline, we construct a multi-stage baseline by adapting GFSLT-VLP \citep{gfslt} (trained originally on pre-segmented clips) to windowed streaming-style supervision. Specifically, we train a GFSLT-VLP on fixed-length windows, using concatenated subtitles of all sentences within each window as the target text, and we replace its original RGB visual encoder with our pose-embedding backbone to ensure a fair comparison of visual inputs. At inference time, we first localize sentences using the best \textbf{StreamSLST} segmentation head, then feed each localized sentence to the window-trained GFSLT-VLP by padding it to the window size using last-frame repetition. 
% \textbf{StreamSLST} and this cascade therefore translate the same intervals and share one localization result by construction. The cascade keeps the visual-language pretraining of GFSLT-VLP and uses the same three-layer mBART decoder as \textbf{StreamSLST}. 
This baseline reflects a conventional ``localize-then-translate'' design, while controlling for visual representation and testing whether joint optimization is necessary.

\subsection{Main Results}
\label{sec:results}

We organize the discussion of Table~\ref{tab:loc_cap_bobsl_h2s} and Table~\ref{tab:loc_cap_phoenix_csl} around four observations that hold across four streaming benchmarks.

\textit{(i) Joint optimization yields the largest gains under noisy boundary supervision.}
The Multi-Stage cascaded baseline uses the same localized segments as \textbf{StreamSLST}, but translates them with a separately trained translator. 
It nearly matches our joint model on the cleaner How2Sign setting with CSV-realigned sentence timing (B-4: 1.80 vs.~2.00), but shows a much larger gap on BOBSL, where subtitles are automatically aligned and boundaries are noisier (B-4: 0.02 vs.~0.23, approximately $10\times$). 
The controlled synthetic streams show an intermediate gap (Phoenix: 5.86 vs.~8.16; CSL: 1.07 vs.~1.64). 
This suggests that joint training is particularly valuable when localization is imperfect, as the translator can adapt to boundary errors rather than relying on fixed upstream segments. Across four benchmarks and five translation metrics, \textbf{StreamSLST} outperforms the cascaded baseline in 19/20 threshold-averaged comparisons, with METEOR on Streaming-CSL-Daily as the only exception.

\textit{(ii) Strict-tIoU matches translate better.}
Across all datasets, increasing the matching threshold from 0.3 to 0.9 substantially reduces F1 but consistently improves BLEU-4 (Phoenix: 3.83$\to$18.27; CSL: 0.79$\to$3.38; BOBSL: 0.24$\to$0.31; H2S: 1.33$\to$3.46). 
This suggests that high-tIoU predictions are not only better localized but also more semantically coherent: tighter boundaries tend to isolate cleaner signing episodes, whereas looser thresholds admit noisier clips. 

% \textit{(iii) Cross-language comparison should be done with semantic metrics.} Across BSL, ASL, DGS, and CSL, our average BLEURT-RT lies in a narrow band (16.80--34.83), whereas BLEU-4 spans more than an order of magnitude (0.23--8.16). The variation in BLEU-4 is driven primarily by target-language properties -- vocabulary size (Phoenix's restricted weather domain vs.~BOBSL's open broadcast vocabulary), tokenization (Chinese characters in CSL), and the prevalence of paraphrastic vs.~literal subtitles -- not by streaming-task difficulty. We therefore recommend BLEURT and METEOR as the primary cross-language indicators when reporting on \textbf{Streaming SLT}, and treat BLEU-4 as a within-dataset diagnostic.
\textit{(iii) Semantic metrics are more reliable for cross-language comparison.}
Across BSL, ASL, DGS, and CSL, our average BLEURT score remains within a relatively narrow range (16.80--34.83), while BLEU-4 varies by over an order of magnitude (0.23--8.16). 
This large BLEU-4 variation is likely influenced by target-language factors, including vocabulary scope, tokenization, and the degree of paraphrasing in subtitles, rather than streaming difficulty alone. 
We therefore use BLEURT and METEOR as the main cross-language indicators for \textbf{Streaming SLT}, while treating BLEU-4 primarily as a within-dataset diagnostic.

% \textit{(iv) Decoder choice influences localization, not just translation.} mBART beats LSTM in F1 on every dataset (Phoenix: 55.95 vs.~51.35; CSL: 59.37 vs.~46.84; BOBSL: 22.22 vs.~19.11; H2S: 23.84 vs.~21.95), with the largest gap on the synthetic benchmarks where translation supervision is densest. This is noteworthy because the decoder is downstream of the localization head; the only path through which it can move F1 is the shared joint loss back-propagating into the encoder and the boundary MLP. The observation reinforces the synergy story from a complementary angle: a stronger language model not only translates better, it also reshapes the segmentation toward translation-friendly boundaries.
\textit{(iv) Decoder choice affects localization as well as translation.}
mBART outperforms LSTM in F1 across all datasets (Phoenix: 55.95 vs.~51.35; CSL: 59.37 vs.~46.84; BOBSL: 22.22 vs.~19.11; H2S: 23.84 vs.~21.95), with the largest gains on the synthetic benchmarks where translation supervision is densest. 
Since the decoder is downstream of the localization head, these F1 gains suggest that the joint loss back-propagates useful translation signals into the shared encoder and boundary predictor. 
% This provides another indication that stronger language modeling can improve not only translation quality, but also the localization learned under joint optimization.

\textbf{Streaming SLT} extends pre-segmented SLT by additionally requiring sentence-boundary discovery. Gray rows in both tables use oracle sentence boundaries and differ in visual input, task construction, and metric aggregation. We therefore present them as reference points rather than directly comparable system rankings. When we rerun the published GFSLT-VLP recipe on pre-segmented Phoenix-2014T clips with our pose front-end in place of its RGB front-end, the result is 4.72 BLEU-4 against the published 21.44, which identifies the visual front-end as a major contributor to the gap to these systems under the pre-segmented protocol itself, before any boundary must be discovered (Appendix~\ref{app:frontend}).

{
\renewcommand{\arraystretch}{1.0}
\begin{table}[t]
\centering
\caption{Localization and Translation performance on the test set of BOBSL (left) and How2Sign (right). The table is divided into \textbf{Streaming SLT} (top) and \textbf{Pre-Segmented SLT} (bottom, in gray). Gray rows are published systems that use RGB input and oracle sentence boundaries under corpus-level metrics, listed as reference points. The BOBSL scores of GFSLT-VLP and Sign2GPT come from the re-training reported by \cite{litfic}.}
\label{tab:loc_cap_bobsl_h2s}
\tiny
\resizebox{\textwidth}{!}{
\begin{tabular}{l | c | c c c c c c | c c c c c c}
\toprule

\multirow{3}{*}{\textbf{Method}} & 
\multirow{3}{*}{\textbf{tIoU}} & 
\multicolumn{6}{c|}{\textbf{BOBSL}} & 
\multicolumn{6}{c}{\textbf{How2Sign}} \\

\cmidrule(lr){3-8} \cmidrule(lr){9-14}

& & 
\textbf{Loc.} & \multicolumn{5}{c|}{\textbf{Translation}} & 
\textbf{Loc.} & \multicolumn{5}{c}{\textbf{Translation}} \\

\cmidrule(lr){3-3} \cmidrule(lr){4-8} \cmidrule(lr){9-9} \cmidrule(lr){10-14}

& & 
\textbf{F1} & \textbf{B-4} & \textbf{B-RT} & \textbf{R-L} & \textbf{CIDEr} & \textbf{Meteor} &
\textbf{F1} & \textbf{B-4} & \textbf{B-RT} & \textbf{R-L} & \textbf{CIDEr} & \textbf{Meteor} \\

\midrule
\multicolumn{14}{c}{\textit{Streaming Sign Language Translation}} \\
\midrule

% -------------- mBART --------------
\multirow{5}{*}{\begin{tabular}{l}
StreamSLST \\ $[$mBART$]$
\end{tabular}}
& 0.3 & 45.66 & 0.24 & 33.91 & 7.31 & 4.48 & 9.06 & 52.37 & 1.33 & 28.24 & 12.49 & 1.00 & 14.85 \\
& 0.5 & 26.73 & 0.20 & 34.97 & 7.38 & 4.90 & 9.52 & 27.84 & 1.48 & 27.84 & 12.48 & 1.47 & 14.21 \\
& 0.7 & 8.69 & 0.18 & 35.32 & 7.37 & 5.48 & 9.50 &  8.86 & 1.73 & 27.55 & 11.11 & 2.02 & 13.49 \\
& 0.9 & 0.92 & 0.31 & 35.12 & 7.73 & 6.37 & 9.47 &  1.44 & 3.46 & 31.62 & 14.75 & 3.06 & 22.43 \\
\cmidrule(lr){2-14}
& Avg. & \textbf{22.22} & \textbf{0.23} & \textbf{34.83} & 7.45 & \textbf{5.31} & \textbf{9.39} & \textbf{23.84} & \textbf{2.00} & 28.81 & \textbf{12.71} & \textbf{1.89} & \textbf{16.24} \\

\cmidrule(r){1-1} \cmidrule(lr){2-14}

% -------------- LSTM --------------
\multirow{5}{*}{\begin{tabular}{l}
StreamSLST \\ $[$LSTM$]$
\end{tabular}}
& 0.3 & 41.05 & 0.05 & 14.75 & 9.26 & 2.98 & 7.93 & 46.92 & 1.05 & 32.39 &  9.64 & 0.31 &  8.15 \\
& 0.5 & 21.64 & 0.06 & 14.90 & 10.15 & 3.31 & 8.50 & 26.75 & 1.45 & 33.27 &  9.19 & 0.40 &  8.80 \\
& 0.7 & 8.33 & 0.08 & 14.69 & 10.21 & 3.69 & 8.58 & 11.17 & 1.87 & 32.06 &  8.36 & 0.34 &  7.90 \\
& 0.9 & 0.78 & 0.13 & 16.24 & 10.37 & 3.58 & 8.80 &  1.32 & 2.93 & 32.17 &  9.64 & 0.71 &  9.20 \\
\cmidrule(lr){2-14}
& Avg. & 19.11 & 0.08 & 15.14 & \textbf{10.00} & 3.39 & 8.45 & 21.95 & 1.83 & \textbf{32.47} & 9.21 & 0.44 & 8.51 \\

\cmidrule(r){1-1} \cmidrule(lr){2-14}

% -------------- Cascaded --------------
\multirow{5}{*}{\begin{tabular}{l}
Multi-Stage \\ (Cascaded)
\end{tabular}}
& 0.3 & 45.66 & 0.02 & 4.12 & 2.45 & 0.55 & 2.15 & 52.37 & 1.62 & 28.11 & 10.32 & 0.69 & 10.86 \\
& 0.5 & 26.73 & 0.02 & 4.15 & 2.50 & 0.58 & 2.18 & 27.84 & 1.82 & 28.44 & 10.01 & 0.74 & 10.41 \\
& 0.7 & 8.69 & 0.01 & 4.22 & 2.62 & 0.61 & 2.24 & 8.86 & 1.92 & 27.99 & 10.65 & 0.83 & 10.67 \\
& 0.9 & 0.92 & 0.01 & 4.30 & 2.75 & 0.65 & 2.31 & 1.44 & 1.82 & 27.78 & 12.35 & 0.31 &  9.52 \\
\cmidrule(lr){2-14}
& Avg. & \textbf{22.22} & 0.02 & 4.20 & 2.58 & 0.60 & 2.22 & \textbf{23.84} & 1.80 & 28.08 & 10.83 & 0.64 & 10.36 \\

\midrule
\multicolumn{14}{c}{\textit{Pre-Segmented Sign Language Translation (oracle boundaries, reference only)}} \\
\midrule

% -------------- Pre-Segmented --------------
\color{gray}\cite{gfslt} GFSLT-VLP & & \color{gray}-- & \color{gray}0.6 & \color{gray}27.7 & \color{gray}7.4 & \color{gray}4.3 & \color{gray}-- & \color{gray}-- & \color{gray}-- & \color{gray}-- & \color{gray}-- & \color{gray}-- & \color{gray}-- \\
\color{gray}\cite{sign2gpt} Sign2GPT & & \color{gray}-- & \color{gray}0.7 & \color{gray}34.3 & \color{gray}10.6 & \color{gray}12.8 & \color{gray}-- & \color{gray}-- & \color{gray}-- & \color{gray}-- & \color{gray}-- & \color{gray}-- & \color{gray}-- \\
\color{gray}\cite{sign2gpt} Sign2GPT w/PGP & & \color{gray}-- & \color{gray}0.9 & \color{gray}35.2 & \color{gray}11.4 & \color{gray}16.1 & \color{gray}-- & \color{gray}-- & \color{gray}-- & \color{gray}-- & \color{gray}-- & \color{gray}-- & \color{gray}-- \\
\color{gray}\cite{litfic} LiTFiC (Video Only) & & \color{gray}-- & \color{gray}2.6 & \color{gray}37.8 & \color{gray}15.6 & \color{gray}37.5 & \color{gray}-- & \color{gray}-- & \color{gray}11.8 & \color{gray}44.1 & \color{gray}31.1 & \color{gray}93.3 & \color{gray}-- \\

\bottomrule
\end{tabular}
}
\end{table}
}

{
\renewcommand{\arraystretch}{1.0}
\begin{table}[t]
\centering
\caption{Localization and Translation performance on the synthetic test set of Streaming-Phoenix-2014T (left) and Streaming-CSL-Daily (right). Gray rows follow Table~\ref{tab:loc_cap_bobsl_h2s} and list published systems on the original Phoenix-2014T and CSL-Daily.}
\label{tab:loc_cap_phoenix_csl}
\tiny
\resizebox{\textwidth}{!}{
\begin{tabular}{l | c | c c c c c c | c c c c c c}
\toprule

\multirow{3}{*}{\textbf{Method}} & 
\multirow{3}{*}{\textbf{tIoU}} & 
\multicolumn{6}{c|}{\textbf{Streaming-Phoenix-2014T}} & 
\multicolumn{6}{c}{\textbf{Streaming-CSL-Daily}} \\

\cmidrule(lr){3-8} \cmidrule(lr){9-14}

& & 
\textbf{Loc.} & \multicolumn{5}{c|}{\textbf{Translation}} & 
\textbf{Loc.} & \multicolumn{5}{c}{\textbf{Translation}} \\

\cmidrule(lr){3-3} \cmidrule(lr){4-8} \cmidrule(lr){9-9} \cmidrule(lr){10-14}

& & 
\textbf{F1} & \textbf{B-4} & \textbf{B-RT} & \textbf{R-L} & \textbf{CIDEr} & \textbf{Meteor} &
\textbf{F1} & \textbf{B-4} & \textbf{B-RT} & \textbf{R-L} & \textbf{CIDEr} & \textbf{Meteor} \\

\midrule
\multicolumn{14}{c}{\textit{Streaming Sign Language Translation}} \\
\midrule

% -------------- mBART --------------
\multirow{5}{*}{\begin{tabular}{l}
StreamSLST \\ $[$mBART$]$
\end{tabular}}
& 0.3 & 83.14 &  3.83 & 28.03 & 11.41 &  2.98 & 12.66 & 86.64 & 0.79 & 15.97 & 12.38 & 0.92 &  6.91 \\
& 0.5 & 77.18 &  4.68 & 28.12 & 11.93 &  3.90 & 13.44 & 73.32 & 1.06 & 16.33 & 12.11 & 1.06 &  6.66 \\
& 0.7 & 56.71 &  5.87 & 29.97 & 13.98 &  5.33 & 14.82 & 55.55 & 1.33 & 17.42 & 11.74 & 1.26 &  6.84 \\
& 0.9 &  6.39 & 18.27 & 41.37 & 29.37 & 19.36 & 28.28 & 14.71 & 3.38 & 17.50 & 14.14 & 2.29 &  8.94 \\
\cmidrule(lr){2-14}
& Avg. & \textbf{55.95} & \textbf{8.16} & \textbf{31.87} & \textbf{16.68} & \textbf{7.89} & \textbf{17.30} & \textbf{59.37} & \textbf{1.64} & \textbf{16.80} & \textbf{12.59} & \textbf{1.38} & 7.34 \\

\cmidrule(r){1-1} \cmidrule(lr){2-14}

% -------------- LSTM --------------
\multirow{5}{*}{\begin{tabular}{l}
StreamSLST \\ $[$LSTM$]$
\end{tabular}}
& 0.3 & 86.94 & 1.26 &  3.71 &  9.93 & 0.61 &  4.18 & 68.62 & 0.56 & 2.35 & 10.20 & 0.16 & 5.53 \\
& 0.5 & 77.60 & 1.36 &  4.08 & 10.62 & 0.74 &  4.53 & 54.61 & 0.74 & 2.50 &  9.97 & 0.20 & 6.11 \\
& 0.7 & 39.03 & 1.47 &  5.78 & 11.66 & 0.83 &  5.14 & 41.26 & 1.31 & 3.81 &  9.67 & 0.70 & 5.84 \\
& 0.9 &  1.82 & 0.67 &  0.54 & 20.69 & 0.09 &  2.56 & 21.72 & 2.12 & 3.77 & 11.84 & 1.30 & 6.97 \\
\cmidrule(lr){2-14}
& Avg. & 51.35 & 1.19 &  3.53 & 13.22 & 0.57 &  4.10 & 46.84 & 1.18 & 3.11 & 10.42 & 0.59 & 6.11 \\

\cmidrule(r){1-1} \cmidrule(lr){2-14}

% -------------- Cascaded --------------
\multirow{5}{*}{\begin{tabular}{l}
Multi-Stage \\ (Cascaded)
\end{tabular}}
& 0.3 & 83.14 &  3.55 & 16.91 &  9.57 & 1.39 & 10.90 & 86.64 & 0.95 & 14.16 & 10.73 & 0.29 & 9.42 \\
& 0.5 & 77.18 &  3.77 & 16.68 &  9.96 & 1.40 & 11.44 & 73.32 & 0.96 & 14.18 & 10.72 & 0.30 & 9.42 \\
& 0.7 & 56.71 &  4.44 & 16.23 & 10.05 & 2.14 & 11.64 & 55.55 & 1.01 & 14.20 & 10.73 & 0.33 & 9.44 \\
& 0.9 &  6.39 & 11.69 & 20.03 & 22.91 & 6.23 & 26.20 & 14.71 & 1.35 & 14.62 & 10.84 & 0.46 & 9.55 \\
\cmidrule(lr){2-14}
& Avg. & \textbf{55.95} & 5.86 & 17.46 & 13.12 & 2.79 & 15.05 & \textbf{59.37} & 1.07 & 14.29 & 10.76 & 0.35 & \textbf{9.46} \\

\midrule
\multicolumn{14}{c}{\textit{Pre-Segmented Sign Language Translation (oracle boundaries, reference only)}} \\
\midrule

% -------------- Pre-Segmented --------------
\color{gray}\cite{gfslt} GFSLT-VLP & & \color{gray}-- & \color{gray}21.44 & \color{gray}-- & \color{gray}42.49 & \color{gray}-- & \color{gray}-- & \color{gray}-- & \color{gray}11.00 & \color{gray}-- & \color{gray}36.44 & \color{gray}-- & \color{gray}-- \\
\color{gray}\cite{sign2gpt} Sign2GPT & & \color{gray}-- & \color{gray}19.42 & \color{gray}-- & \color{gray}45.23 & \color{gray}-- & \color{gray}-- & \color{gray}-- & \color{gray}12.96 & \color{gray}-- & \color{gray}41.12 & \color{gray}-- & \color{gray}-- \\
\color{gray}\cite{sign2gpt} Sign2GPT w/PGP & & \color{gray}-- & \color{gray}22.52 & \color{gray}-- & \color{gray}48.90 & \color{gray}-- & \color{gray}-- & \color{gray}-- & \color{gray}15.40 & \color{gray}-- & \color{gray}42.36 & \color{gray}-- & \color{gray}-- \\

\bottomrule
\end{tabular}
}
\end{table}
}

\subsection{Ablation Studies}

\label{sec:ablation}

We isolate three design decisions and the synergy effect on Streaming-Phoenix-2014T and How2Sign.

\textbf{Component Ablations.} Table~\ref{tab:ablation1} examines three components. \textit{First} ablation removes the localized visual features and conditions translation on the query alone. Localization stays close to the full model at 54.23 and 24.08 F1, while BLEU-4 falls from 8.16 to 2.89 on Streaming-Phoenix-2014T and from 2.00 to 1.51 on How2Sign. The query is therefore sufficient to locate a sentence, and the localized encoder features supply the detail needed to translate it.
\textit{Second} ablation keeps the full model and its input unchanged and stops the translation loss from updating the shared queries, encoder, and backbone. Translation falls on both benchmarks, with BLEU-4 dropping from 8.16 to 5.81 and from 2.00 to 1.24. Localization also changes in opposite directions on two benchmarks, so we interpret the result as end-to-end coupling between two objectives through the shared representation. This control complements the cascade comparison in Section~\ref{sec:results}, since it holds the architecture and input fixed and removes only the shared gradient.
\textit{Third} ablation removes the Stage-1 pretraining. Translation falls on both benchmarks, with BLEU-4 dropping from 8.16 to 3.65 and from 2.00 to 0.95. On Streaming-Phoenix-2014T, localization also falls from 55.95 to 29.90 F1, and Appendix~\ref{sec:collapse_analysis} traces this drop to a representation collapse.
This result also extends the insight of \cite{gfslt} -- that language-guided representation learning is important for gloss-free SLT -- to the streaming setting, where the same representation must support both boundary discovery and translation generation.

% isolates the effect of VLP within the same \textbf{StreamSLST} architecture. Training the joint model from random initialization leads to substantially worse translation performance. It struggles to align pose features with linguistic semantics under noisy window-level supervision.
% Adding our tri-modal contrastive pretraining (Stage 1 + Stage 2) improves all translation metrics while leaving localization largely unchanged.
% This suggests that pretraining mainly strengthens language-aligned pose representations rather than the eventness detector.

\begin{table}[t]
\centering
\caption{Component ablations of \textbf{StreamSLST} on Streaming-Phoenix-2014T (left) and How2Sign (right). Each variant keeps the architecture, data, and schedule of the full model and changes the component named in its row. All reported metrics are averaged across all tIoU thresholds.}
\label{tab:ablation1}
\tiny
\resizebox{\textwidth}{!}{
\begin{tabular}{l | c c c c c | c c c c c}
\toprule

\multirow{2}{*}{\textbf{Variant}} &
\multicolumn{5}{c|}{\textbf{Streaming-Phoenix-2014T}} &
\multicolumn{5}{c}{\textbf{How2Sign}} \\

\cmidrule(lr){2-6} \cmidrule(lr){7-11}

& \textbf{F1} & \textbf{B-4} & \textbf{B-RT} & \textbf{R-L} & \textbf{Meteor} &
  \textbf{F1} & \textbf{B-4} & \textbf{B-RT} & \textbf{R-L} & \textbf{Meteor} \\

\midrule

\textbf{StreamSLST} (full) & 55.95 & \textbf{8.16} & \textbf{31.87} & \textbf{16.68} & \textbf{17.30} & 23.84 & \textbf{2.00} & 28.81 & \textbf{12.71} & \textbf{16.24} \\
\xmark\, localized visual conditioning & 54.23 & 2.89 & 31.33 & 10.40 & 10.38 & 24.08 & 1.51 & \textbf{31.15} & 10.67 & 13.96 \\
\xmark\, translation gradient & 51.86 & 5.81 & 26.32 & 10.66 & 11.93 & 26.25 & 1.24 & 27.00 & 8.72 & 12.99 \\
\xmark\, Stage-1 pretraining & 29.90 & 3.65 & 19.97 & 10.59 & 10.26 & 24.85 & 0.95 & 24.29 & 10.13 & 10.90 \\

\bottomrule
\end{tabular}
}
\end{table}

\textbf{Synergistic Relationship.} To analyze the interaction between localization and translation, we compare our full model with an oracle-boundary variant that uses annotated sentence boundaries during training and evaluation. As shown in Table~\ref{tab:ablation2}, the full model with learned proposals outperforms the annotated-span variant across threshold-averaged metrics (first row).
We found that learned proposals are wider than annotated spans on average (Appendix~\ref{app:support}), so we further test whether wider oracle support reproduces the learned-proposal results. We retrain the oracle-boundary variant with spans expanded by 0.5\,s and by 1.0\,s on each side, and with spans rescaled by factors of 1.40 and 2.25, corresponding to the mean learned-to-annotated length ratio on Streaming-Phoenix-2014T and the median length ratio on How2Sign. Each variant is trained and evaluated on its own support. Under the reported evaluation protocol, learned proposals obtain higher translation scores than all tested oracle-support variants on both benchmarks. Joint optimization therefore learns wider, translation-effective temporal support that simple expansion or rescaling of annotated spans fails to reproduce. Together with the shared-gradient ablation, this motivates studying localization and translation jointly, while leaving the specific cause of the temporal-support advantage open.

\begin{table}[t]
\centering
\caption{Learned proposals vs. Oracle-support controlled proposals on Streaming-Phoenix-2014T (left) and How2Sign (right). All reported metrics are averaged across all tIoU thresholds.
}
\label{tab:ablation2}
\small
\resizebox{0.9\textwidth}{!}{
\begin{tabular}{l | c c c c | c c c c}
\toprule
\multirow{2}{*}{\textbf{Temporal support}} & \multicolumn{4}{c|}{\textbf{Streaming-Phoenix-2014T}} & \multicolumn{4}{c}{\textbf{How2Sign}} \\
\cmidrule(lr){2-5} \cmidrule(lr){6-9}
 & \textbf{B-4} & \textbf{B-RT} & \textbf{R-L} & \textbf{Meteor} & \textbf{B-4} & \textbf{B-RT} & \textbf{R-L} & \textbf{Meteor} \\
\midrule
GT proposals & 3.08 & 26.50 & 12.29 & 12.02 & 1.51 & 27.56 & 11.25 & 14.82 \\
GT $\pm0.5$\,s & 3.79 & 26.27 & 12.14 & 13.41 & 1.00 & 24.72 & 11.65 & 13.43 \\
GT $\pm1.0$\,s & 3.26 & 26.41 & 9.09 & 11.05 & 1.27 & 28.37 & 12.37 & 14.80 \\
GT $\times1.40$ & 2.58 & 21.93 & 9.47 & 10.77 & 1.50 & 27.97 & 11.41 & 14.35 \\
GT $\times2.25$ & 3.09 & 30.13 & 6.14 & 8.32 & 1.17 & 27.38 & 11.99 & 14.28 \\
\midrule
Learned proposals & \textbf{8.16} & \textbf{31.87} & \textbf{16.68} & \textbf{17.30} & \textbf{2.00} & \textbf{28.81} & \textbf{12.71} & \textbf{16.24} \\
\bottomrule
\end{tabular}
}
\end{table}

\section{Conclusion}
In this work, we have studied the shift from pre-segmented clip-based translation to \textbf{Streaming Sign Language Translation (Streaming SLT)}, a realistic paradigm that unifies temporal segmentation and translation within continuous, untrimmed video streams. 
To tackle the inherent challenges, we present \textbf{StreamSLST}, a unified framework tackling the \textbf{Streaming SLT} task via a Deformable Transformer. 
By integrating a \textit{Sentence-Aware Sliding Window strategy} and \textit{Tri-modal Pretraining}, we successfully scale our approach to long-form, unsegmented videos. 
Experiments demonstrate that joint optimization learns wider, translation-effective temporal support.
Experiments show that the joint model improves translation relative to the matched cascade under and our temporal-support controls further characterize learned proposal behavior.
Our unified approach establishes a solid foundation for future real-time sign language interfaces.

\section*{Acknowledgement}
This research is funded in part by ARC-Discovery grant (DP220100800 to XY), ARC-DECRA grant (DE230100477 to XY, DE260100509 to Yanbin) and Google Research Scholar Program. 
% Yanbin is funded by ARC-DECRA grant (DE260100509). 
We also gratefully thank all the anonymous reviewers and ACs for their constructive comments.

% The file named.bst is a bibliography style file for BibTeX 0.99c
\clearpage

\bibliographystyle{ref_style}
\bibliography{neurips_2026}

%%%%%%%%%%%%%%%%%%%%%%%%%%%%%%%%%%%%%%%%%%%%%%%%%%%%%%%%%%%%

\clearpage
\appendix
\section*{Appendix}
\noindent This appendix is organized as follows:
\begin{multicols}{2}
\begin{itemize}[leftmargin=15pt]
\item Comparison with Related Task Settings (Section~\ref{app:positioning})
\item Dataset Details (Section~\ref{app:datasets})
\item Data Pre-processing (Section~\ref{sec:data_preprocessing})
\item Further Setup Details (Section~\ref{sec:further_setup})
\item Representation Collapse Analysis (Section~\ref{sec:collapse_analysis})
\item Learned and Oracle Temporal Support (Section~\ref{app:support})
\item Visual Front-End Control (Section~\ref{app:frontend})
\item Case Studies (Section~\ref{sec:case_study})
\item Limitations and Future Work (Section~\ref{sec:limitations_future})
\end{itemize}
\end{multicols}

\section{Comparison with Related Task Settings}
\label{app:positioning}

Table~\ref{tab:positioning} places \textbf{Streaming SLT} among the settings of the related work by what a system receives at test time and what it returns. Alongside Pre-segmented SLT translates one clip whose boundaries are given, three lines of work address temporal structure, yet still leave the practical case uncovered. They differ from \textbf{Streaming SLT} in two respects that determines everything else: what each system receives at test time and what it returns. (1) \textit{Segmentation} methods receives video and return boundaries of signs, phrases, or subtitle units without producing text~\citep{bull2020,renz2021,bioseg,rao-etal-2025-sign}, while (2) \textit{Alignment} work receives video together with the target text and returns timings for that text~\citep{bull2021,sea2026}. (3) \emph{Incremental SLT} shortens the delay inside one utterance, with recent work consuming a clip whose extent the dataset supplies and still depending on gloss annotation~\citep{simulslt,zuo2024towards}. \textbf{Streaming SLT} receives video alone and asks for the extent and text together. Our contribution is therefore the joint discovery and gloss-free translation of sentence events in untrimmed multi-sentence streams, together with the translation head and the evaluation protocol that this task requires.

\begin{table}[h]
\centering
\caption{Where \textbf{Streaming SLT} sits among related settings.}
\label{tab:positioning}
\footnotesize
\resizebox{0.99\textwidth}{!}{
\begin{tabular}{l l l c l}
\toprule
\textbf{Setting} & \textbf{Input at test time} & \textbf{Sentence extent} & \textbf{Gloss-free} & \textbf{Output} \\
\midrule
Pre-segmented SLT~\citep{gfslt,sign2gpt,unisign,litfic,scalingslt} & sentence clip & given & yes & text \\
Simultaneous SLT~\citep{simulslt} & sentence clip & given & no & incremental text \\
Online SLT~\citep{zuo2024towards} & continuous signing & sign-level windows & no & text via glosses \\
Segmentation~\citep{bull2020,renz2021,bioseg,rao-etal-2025-sign} & untrimmed video & predicted & -- & boundaries \\
Subtitle alignment~\citep{bull2021,sea2026} & video and subtitle text & aligned to the text & -- & timings \\
\midrule
\textbf{Streaming SLT} (ours) & untrimmed video & predicted & yes & timed text \\
\bottomrule
\end{tabular}
}
\end{table}

\section{Dataset Details}
\label{app:datasets}

We use two types of datasets in our experiments. 
BOBSL and How2Sign are used as real long-form streaming benchmarks, while Streaming-Phoenix-2014T and Streaming-CSL-Daily are synthetic streaming benchmarks constructed from sentence-level sign language translation datasets. 
The real datasets evaluate performance under realistic long-video conditions, whereas the synthetic datasets allow controlled evaluation with exact oracle sentence boundaries.

\subsection{Real Long-form Datasets}
\label{app:real_datasets}

\paragraph{BOBSL.}
BOBSL~\citep{bobsl} is a large-scale British Sign Language dataset collected from BBC broadcast videos. The videos are long-form and unconstrained, often containing signed utterances, idle intervals, scene changes, and subtitle misalignment. The release ships two subtitle-alignment subsets: an \emph{auto-aligned} set covering all $\sim$1.4K hours, with timing produced by an automatic subtitle aligner~\citep{bull2021} whose noise is itself the subject of recent work~\citep{rao-etal-2025-sign,sea2026}, and a \emph{manually-aligned} subset of 55 episodes with human-verified timing, 35 of which belong to our test split. We train on the auto-aligned subset (large but noisy) under our sliding-window sampler and report streaming results on its held-out test split. On 35 test episodes with manually-aligned subtitles, the automatic timings match the human timings with F1 of 74.98, 61.89, 40.22, and 10.80 at tIoU 0.3, 0.5, 0.7, and 0.9, and with an average F1 of 46.98 under our protocol (Appendix~\ref{app:metrics}). BOBSL references thus become unreliable at strict thresholds, and localization on this corpus is most informative at tIoU 0.5 and below. All streaming systems in Table~\ref{tab:loc_cap_bobsl_h2s} are trained and evaluated on the same automatic boundaries, so their comparison stays controlled. 
% The gap-duration distribution from the manually-aligned subset is used to guide the data construction of \textbf{Streaming-Phoenix-2014T} and \textbf{Streaming-CSL-Daily} by simulating realistic streaming settings.
We use gap durations of the manual alignment to guide the pause distribution of two synthetic benchmarks to simulate realistic streaming settings.

\paragraph{How2Sign.}
How2Sign~\cite{duarte2021how2sign} is a continuous American Sign Language dataset consisting of instructional videos paired with English transcripts and additional modalities. 
Compared with BOBSL, it differs in signing language, domain, visual style, and recording setup. 
% We include How2Sign to evaluate whether our streaming framework generalizes beyond British Sign Language broadcast videos.
Although the source videos are real continuous footage, the released annotations are organized as per-sentence clips with separately-extracted OpenPose ~\citep{cao2017realtime} keypoints and a CSV of realigned per-sentence timestamps in the source-video timeline. To make the dataset usable in our pose-stream framework, we reassemble each video as a single continuous pose stream: we (1) convert OpenPose-137 keypoints into the COCO-WholeBody-133~\citep{cocobody} layout used by the rest of our pipeline, (2) place each sentence at its CSV-realigned start frame on the original video timeline, (3) bridge inter-sentence gaps with linear ($C^0$) interpolation, and (4) trim leading/trailing rest frames to suppress the trivial ``hands-down = boundary'' cue. Unlike \textit{Streaming-Phoenix-2014T} and \textit{Streaming-CSL-Daily}, no inter-sentence pause distribution is sampled -- the inter-sentence silences are the real gaps from the source video. We treat How2Sign as a real-world streaming dataset because both the inter-sentence timing and the per-sentence pose dynamics originate from genuine continuous ASL signing; only the format conversion is synthetic. Implementation details are in our public release.

\paragraph{Preprocessing and window sampling.}
For both BOBSL and How2Sign, we treat each video as a continuous stream rather than a collection of isolated sentence clips. 
We extract pose sequences and align them with the available subtitle or transcript timestamps. 
During training, long videos are divided into fixed-length sliding windows. 
For a window spanning $[a,b]$, the target text is obtained by collecting all text segments that lie inside the window and last between 1 and 20 seconds:
\begin{equation}
\mathcal{Y}_{[a,b]} =
\{\, y_i \mid a \le t_i^{\mathrm{start}},\; t_i^{\mathrm{end}} \le b,\; 1 \le t_i^{\mathrm{end}} - t_i^{\mathrm{start}} \le 20 \,\}.
\end{equation}
The selected text segments are concatenated in temporal order. 
This protocol allows training on long-form videos without requiring perfect sentence-level segmentation.

\subsection{Synthetic Streaming Benchmarks}
\label{app:synthetic_datasets}

% To evaluate streaming translation under controlled boundary supervision, we construct two synthetic streaming benchmarks from sentence-level datasets: \textbf{Streaming-Phoenix-2014T} and \textbf{Streaming-CSL-Daily}. 
% Phoenix-2014T~\citep{camgoz2018neural} is a German Sign Language translation benchmark, while CSL-Daily~\citep{csldaily} is a Chinese Sign Language translation dataset covering daily-life topics. 

To evaluate streaming translation under controlled boundary supervision, we construct two synthetic streaming benchmarks, \textbf{Streaming-Phoenix-2014T} and \textbf{Streaming-CSL-Daily}, from the sentence-level corpora Phoenix-2014T~\citep{camgoz2018neural} and CSL-Daily~\citep{csldaily}. Phoenix-2014T~\citep{camgoz2018neural} is a German Sign Language translation benchmark, while CSL-Daily~\citep{csldaily} is a Chinese Sign Language translation dataset covering daily-life topics. Each source corpus provides per-sentence clips paired with text but no continuous-stream timeline; we therefore reassemble multiple clips into one continuous pose stream while preserving exact frame-level sentence boundaries by construction.

\paragraph{Stream construction.} Each source sample is $x_i = (P_i, y_i, s_i)$, where $P_i$ is the pose sequence, $y_i$ the text, and $s_i$ the signer identity. For each output stream, we (i) sample a signer whose clip pool $\mathcal{C}_s$ has at least two usable clips ($|\mathcal{C}_s| \geq 2$, restricting the synthesis to multi-clip signers); (ii) sample the number of sentences $K$ uniformly from a fixed range so that streams span multiple model windows; and (iii) select $K$ clips uniformly without replacement from $\mathcal{C}_s$. Same-signer sampling avoids artificial discontinuities due to body-scale, camera-position, or signing-style changes that would otherwise expose the boundary trivially. Each selected clip undergoes the following pipeline before concatenation:
\begin{enumerate}[leftmargin=*]
\item Resample to the per-dataset target frame rate $f_\text{target}$ (12.5 fps for Phoenix, 15 fps for CSL) using linear interpolation on $(x, y)$ coordinates and nearest-neighbor on confidence scores.
\item Trim contiguous leading/trailing rest frames -- frames in which both wrist keypoints lie below the shoulder line in image coordinates. A safeguard floor preserves at least three frames per clip.
\item Drop the clip if the post-trim duration is shorter than 1 second, matching the loader's minimum subtitle duration.
\end{enumerate}
After this pipeline produces $K' \leq K$ usable clips, the stream is assembled as
\begin{equation}
S = B_\mathrm{pre} \;\Vert\; c_1 \;\Vert\; \Delta_1 \;\Vert\; c_2 \;\Vert\; \cdots \;\Vert\; \Delta_{K'-1} \;\Vert\; c_{K'} \;\Vert\; B_\mathrm{post},
\end{equation}
where every $\Delta_i$ and the two boundary segments $B_\mathrm{pre}, B_\mathrm{post}$ are Hermite-interpolated bridges (described below). Only the $c_i$ are added to the transcript; bridges are non-transcribed.

\paragraph{Inter-sentence bridges.} Real long-form sign-language streams contain co-articulated transitions: the signer rarely returns to a full rest pose between adjacent sentences. We approximate this by sampling each inter-sentence gap from an empirical distribution $\mathcal{G}$ derived from BOBSL manual-aligned annotations:
\begin{equation}
g_i \sim \mathcal{G}, \qquad \mathcal{G} = \big\{ \max(0, \, t_{j+1}^\mathrm{start} - t_j^\mathrm{end}) \big\}_{j} \text{ over BOBSL subtitles,}
\end{equation}
of which approximately 74\% of values are exactly zero (co-articulated boundaries) and the remaining 26\% form a positive right-tailed distribution (visible pauses). The gap is converted to a frame count
\begin{equation}
L_i = \max\bigl(L_\text{min},\ \mathrm{round}(g_i \cdot f_\text{target})\bigr),
\end{equation}
with a minimum bridge length $L_\text{min} = 2$ frames; \emph{a Hermite bridge is always inserted}, even when the sampled gap is zero, so that the seam between adjacent clips remains $C^1$-continuous in pose velocity rather than a hard discontinuity. The bridge interpolates per-keypoint $(x, y)$ coordinates between $c_i[-1]$ (the last frame of clip $i$) and $c_{i+1}[0]$ using cubic Hermite interpolation with endpoint tangents set to the per-frame velocity at each side ($c_i[-1] - c_i[-2]$ and $c_{i+1}[1] - c_{i+1}[0]$). Tangent magnitudes are damped inversely with bridge length and clamped to $\leq 2\|c_{i+1}[0] - c_i[-1]\|$ per joint to prevent overshoot when long bridges propagate clip-end momentum across many frames; confidence scores along the bridge are the elementwise minimum of the two endpoints.

\paragraph{Stream lead-in and lead-out.} $B_\mathrm{pre}$ and $B_\mathrm{post}$ model the silent broadcast lead-in and lead-out characteristic of long-form recordings. Their lengths are sampled from the \emph{positive-only} subset of $\mathcal{G}$ (so $L_\text{pre}, L_\text{post} > 0$ always). Each is itself a Hermite bridge: $B_\mathrm{pre}$ interpolates from the last frame of a randomly chosen \emph{phantom clip} -- a same-signer clip not included in the $K$ selected -- to the first frame of $c_1$; $B_\mathrm{post}$ does the symmetric thing from $c_{K'}$ to a phantom clip's first frame. We use only the phantom's boundary frame as the bridge endpoint (its interior is not inserted), and we deliberately do not trim the phantom's rest frames so that the bridge endpoints carry realistic broadcast-rest pose statistics. If the signer pool has no spare clip outside the chosen $K$, we fall back to using $c_1$ or $c_{K'}$ itself as the phantom; the bridge is still Hermite-interpolated rather than a held frame.

\paragraph{Oracle boundaries.} A running frame counter $r$ is updated as the stream is built; when $c_i$ is appended its oracle boundaries are recorded as $b_i^\mathrm{start} = r,\ b_i^\mathrm{end} = r + |c_i|$, and converted to timestamps via $f_\text{target}$. Bridges and phantom-derived segments are never transcribed, ensuring the final boundary annotations are exact and frame-aligned.

\paragraph{Benchmark statistics.} Table~\ref{tab:benchmark_stats} reports the size of each benchmark that we assemble from sentence-level clips. The manifest records the source clips and the sampled pauses of every stream, and the generator uses a fixed random seed, so the streams can be rebuilt from the source corpora.

\begin{table}[h]
\centering
\caption{Statistics of the benchmarks that we assemble from sentence-level clips. Hours count the full length of each stream, including the segments between sentences.}
\label{tab:benchmark_stats}
\small
\begin{tabular}{l l r r r}
\toprule
\textbf{Benchmark} & \textbf{Split} & \textbf{Streams} & \textbf{Hours} & \textbf{Sentences} \\
\midrule
\multirow{3}{*}{Streaming-Phoenix-2014T} & train & 564 & 6.99 & 4,332 \\
 & val & 27 & 0.10 & 57 \\
 & test & 38 & 0.17 & 95 \\
\midrule
\multirow{3}{*}{Streaming-CSL-Daily} & train & 4,565 & 25.76 & 15,056 \\
 & val & 269 & 1.54 & 880 \\
 & test & 294 & 1.72 & 957 \\
\midrule
\multirow{3}{*}{How2Sign} & train & 2,179 & 43.06 & 30,898 \\
 & val & 115 & 2.26 & 1,732 \\
 & test & 149 & 3.31 & 2,332 \\
\bottomrule
\end{tabular}
\end{table}

\section{Data Pre-processing}
\label{sec:data_preprocessing}

\begin{figure}[h] % Positioning: h = here
    \centering
    \includegraphics[width=0.8\textwidth]{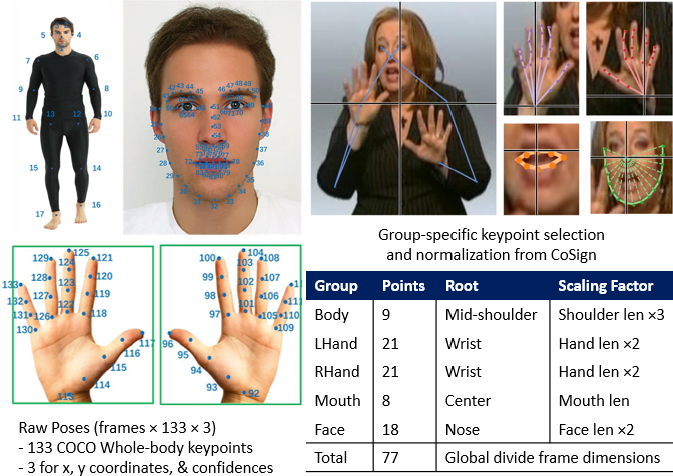}
    \caption{Robust pose representation strategy from CoSign for COCO Whole-body keypoints}
    \label{fig:pose}
\end{figure}

Our preprocessing pipeline addresses 2 critical challenges: extracting robust pose representations, as shown in Figure \ref{fig:pose}, and handling the computational constraints of processing long untrimmed videos, as shown in Figure \ref{fig:sampling}.
We apply a uniform pose-extraction and pre-processing pipeline to all four benchmarks, with only dataset-specific source frame rates and canvas dimensions varying. Source poses are obtained from DWPose~\citep{dwpose} for BOBSL and our synthesized streams, and OpenPose~\citep{cao2017realtime} for How2Sign (re-mapped to the COCO-WholeBody-133 layout, see Appendix~\ref{app:real_datasets}). Low-confidence points (< 0.5) are zeroed to reduce noise, as in \cite{cosign}. 

% This window size is also informed by our analysis of subtitle duration statistics in real streaming datasets like BOBSL, as shown in Figure \ref{fig:bobsl}.

\subsection{Pose Extraction and Normalization}

The 2D skeletal poses extraction process yields 133 COCO WholeBody \citep{cocobody} keypoints per frame. Following CoSign \citep{cosign}, we retain 77 keypoints organized into 5 anatomically-meaningful groups: upper body (9 keypoints), left hand (21), right hand (21), mouth (8 inner keypoints), and lower face (18 keypoints). This selection preserves critical co-occurrence signals between manual and non-manual articulators while reducing computational overhead. Each anatomical group $g$ undergoes group-specific centralization, where keypoints are normalized relative to their local root (e.g., wrist for hands, mid-shoulder for body) and scaled by characteristic anatomical lengths (e.g., shoulder width for body, palm size for hands):
\begin{align}
\mathbf{k}_g^{\text{norm}}(t) = \frac{\mathbf{k}_g(t) - \mathbf{r}_g(t)}{L_g(t) + \epsilon} \cdot s_g,
\end{align}
where \(\mathbf{r}_g(t)\) is the root keypoint, \(L_g(t)\) the characteristic length, \(s_g\) a group-specific scaling factor, and \(\epsilon = 10^{-6}\) to prevent division by 0. Final coordinates are normalized by video dimensions to achieve spatial invariance. Keypoints with confidence below 0.5 are masked to zero, handling tracking failures and occlusions.

\subsection{Sentence-Aware Sliding Window Sampler}

\begin{figure}[h] % Positioning: h = here
    \centering
    \includegraphics[width=1.0\textwidth]{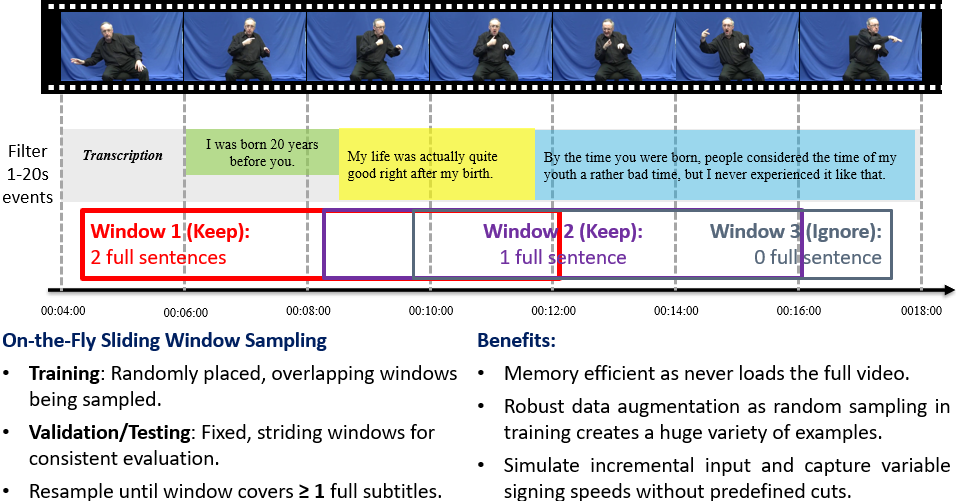}
    \caption{Sliding window sampling for data loading. We dynamically extract training and evaluation samples from long-form videos without manual pre-segmentation. As processing an entire long video is computationally expensive, we divide each video into fixed-size temporal windows of $W$ frames.}
    \label{fig:sampling}
\end{figure}

A key practical challenge in adapting large-scale SLT datasets for \textbf{Streaming SLT} is that the original data can consist of significantly long videos with dozens of sentences. Traditional DVC datasets, like ActivityNet Captions \citep{activitynet}, provide trimmed clips of 3–10 minutes with carefully curated sentence sequences, simplifying the learning problem. In contrast, \textbf{Streaming SLT} requires models to operate on extended temporal contexts where sentence boundaries are distributed irregularly. We address this through a sliding window sampling strategy, as illustrated in Figure \ref{fig:sampling} and Algorithm~\ref{alg:sliding_window}.

\begin{algorithm}[tb]
    \caption{Sentence-Aware Sliding Window Sampling}
    \label{alg:sliding_window}
    \textbf{Input}: split, video list $V$ with frames $T_v$ and subtitle list $S_v$ \\
    \textbf{Parameter}: window size $W$, sentence-count bounds $[M_{\min}, M_{\max}]$, max retry $T_{\max}$ \\
    \textbf{Output}: A window $(v, \text{start}, \text{end})$ containing $M \in [M_{\min}, M_{\max}]$ fully-enclosed valid sentences
    
    \begin{algorithmic}[1] %[1] enables line numbers
        \STATE \textbf{function} CountSentences($v, \text{start}, \text{end}$)
        
        \STATE \quad $C \xleftarrow{} 0$; \textbf{for} each subtitle $s \in S_v$ \textbf{do}
        \STATE \quad \textbf{if} $s$ is valid and fully within $[\text{start}, \text{end}]$ \textbf{then} 
        \STATE \quad \quad $C \xleftarrow{} C+1$
        \STATE \quad \textbf{return} $C$
        
        \STATE
        \STATE \textbf{function} GetItem(index)
        \STATE \quad $v \xleftarrow{} V[\text{index}]$
        \STATE \quad \textbf{if} split is 'train' \textbf{then}
        \STATE \quad \quad $start_{max} \xleftarrow{} \max(0, T_v - W)$
        \STATE \quad \quad \textbf{for} $i=1$ to $T_{\max}$ \textbf{do}
        \STATE \quad \quad \quad $\text{start} \xleftarrow{} \text{RandInt}(0, start_{max})$
        \STATE \quad \quad \quad \textbf{if} $M_{\min} \le$ CountSentences($v, \text{start}, \text{start}+W) \le M_{\max}$ \textbf{then}
        \STATE \quad \quad \quad \quad \textbf{return} $(v, \text{start}, \text{start}+W)$
        % \STATE \quad \quad \quad \textbf{end if}
        % \STATE \quad \quad \textbf{end for}
        \STATE \quad \quad \textbf{return} FallbackWindow($v$) \COMMENT{No window found}
        \STATE \quad \textbf{else} \COMMENT{Use pre-computed fixed windows for eval}
        \STATE \quad \quad \textbf{return} precomputed\_windows[index]
        % \STATE \quad \textbf{end if}

        \STATE
        \STATE \textbf{function} FallbackWindow($v$) \COMMENT{Find densest window}
        \STATE \quad Filter and sort $S_v$ to get valid sentences $E$.
        \STATE \quad Find all sentence sub-sequences in $E$ fitting window $W$.
        \STATE \quad Let $\mathcal{C} \xleftarrow{}$ start ranges of windows with $M_{\min} \le C \le M_{\max}$ sentences.
        \STATE \quad \textbf{if} $\mathcal{C}$ is not empty \textbf{then}
        \STATE \quad \quad $(\text{low}, \text{high}) \xleftarrow{} \text{RandomChoice}(\mathcal{C})$
        \STATE \quad \textbf{else}
        \STATE \quad \quad $(\text{low}, \text{high}) \xleftarrow{}$ start range of the window whose count is closest to $[M_{\min}, M_{\max}]$.
        % \STATE \quad \textbf{end if}
        \STATE \quad $\text{start} \xleftarrow{} \text{RandInt}(\text{low}, \text{high})$
        \STATE \quad \textbf{return} $(v, \text{start}, \text{start}+W)$
    \end{algorithmic}
\end{algorithm}

\subsubsection{Training Sampling: Random Placement with Filtering}

During training, for each video in the training split, we sample fixed-size windows on-the-fly by randomly selecting a start frame $s \sim \text{Uniform}(0, T - W)$, where $T$ is the total video frames, $W$ is the window size, and the corresponding window is $[s, s + W]$. This stochasticity serves as a powerful form of data augmentation, in which multiple ``potentially overlapping'' windows from the same video generate diverse training samples. As a result, it can serve 2 purposes: (1) increase effective training set size via data augmentation, and (2) expose models to varied sentence distributions and temporal contexts every epoch. This process is not only memory-efficient but also introduces variety, thereby preventing overfitting.

However, not all randomly sampled windows contain meaningful training signals. To ensure efficient training, we designed the data loader to impose a minimum sentence constraint: each sampled window must fully contain 1 to 10 complete, valid sentences ($M_{\min} = 1$ and $M_{\max} = N_{\max} = 10$). A sentence/subtitle/event is considered valid if its entire temporal extent (start and end frames) lies within the window boundaries and its duration falls within a specific range of seconds. For example, \cite{litfic} used a duration of $[1.0, 20.0]$ to filter out unusually short or long sentences. If a randomly sampled window contains fewer than 1 valid sentence, we resample up to a maximum number of attempts (e.g., 20). If all attempts fail or no valid window is found after that threshold, we employ a fallback strategy that greedily seeks the densest window (containing most valid sentences) to ensure that training never stalls due to insufficient data. To achieve this, we implemented a two-pointer sweep algorithm: for each potential sentence cluster that fits within the window size, it identifies valid window placements that fully contain all sentences in that cluster. If multiple such placements exist, one is selected uniformly at random. If no placement contains the minimum required sentences, the algorithm falls back to random placement as a safety mechanism.

\subsubsection{Evaluation Sampling: Fixed Stride, Sequential Coverage}

For validation and test splits, we adopt a deterministic, sequential sampling approach to enable consistent evaluation, ensuring fair comparison across runs. Starting from the beginning of each video, we extract windows at a fixed stride equal to a proportion of the window size, potentially producing a certain degree of overlap between consecutive windows. The overlap during evaluation sampling can prevent ``blind spots'' where a sentence can fall between windows and be missed, ensuring complete coverage of the video. Unlike training, we do not resample evaluation windows. Instead, we precompute all valid windows during dataset initialization and store them as a flat list, enabling $O(1)$ access during evaluation. Only windows containing at least 1 valid sentence are retained, ensuring that metrics are computed only on meaningful samples. This fixed-stride approach ensures that evaluation metrics are reproducible and not biased by opportunistic sampling.
The stride is $0.9W$, so consecutive windows overlap by 1.5\,s. Each window is scored on its own, without merging or deduplication across windows, so a sentence is scored twice only when it lies entirely inside this overlap. Table~\ref{tab:duplication} shows that this case is rare and that no sentence is scored in three windows.

\begin{table}[h]
\centering
\caption{Double-counting audit of the evaluation protocol on four test splits. Each sentence is counted once, and the percentages refer to the scored sentences.}
\label{tab:duplication}
\small
\begin{tabular}{l r r r r}
\toprule
\textbf{Test split} & \textbf{Windows} & \textbf{Sentences scored} & \textbf{In two windows} & \textbf{In three windows} \\
\midrule
BOBSL & 43,763 & 98,678 & 1,090 (1.10\%) & 0 \\
How2Sign & 526 & 1,275 & 10 (0.78\%) & 0 \\
Streaming-CSL-Daily & 247 & 544 & 0 (0.00\%) & 0 \\
Streaming-Phoenix-2014T & 22 & 45 & 1 (2.22\%) & 0 \\
\bottomrule
\end{tabular}
\end{table}

\section{Further Setup Details}
\label{sec:further_setup}

\subsection{Window Size and Frame Rate}

To cover longer sequences and reduce computational burden, we extracted poses at half the original fps. We use a single 15-second window across all four benchmarks. We fix the window duration in \emph{seconds}, not frames, so that the same window covers comparable linguistic content across datasets with different acquisition rates. At the per-dataset target frame rate in Table~\ref{tab:dataset_specs}, this corresponds to 187 frames at 12.5 fps (BOBSL, Streaming-Phoenix-2014T) or 225 frames at 15 fps (Streaming-CSL-Daily, How2Sign). Per-dataset source/target rates and canvas sizes are summarized in Table~\ref{tab:dataset_specs}.

\begin{table}[h]
\centering
\small
\caption{Dataset-specific input specifications. All four streams share the same
downstream preprocessing (group-relative pose normalization, sliding-window
sampling, mBART-cc25 tokenization).}
\setlength{\tabcolsep}{4pt}
\label{tab:dataset_specs}
\begin{tabular}{l c c c c l}
\toprule
\textbf{Dataset}
  & \textbf{Src. FPS}
  & \textbf{Tar. FPS}
  & \textbf{Canvas}
  & \textbf{Lang.}
  & \textbf{Type} \\
\midrule
BOBSL
  & 25 & 12.5 & 444$\times$444
  & en\_XX & Real (broadcast) \\
How2Sign
  & 30 & 15 & 1280$\times$720
  & en\_XX & Real (re-streamed) \\
Streaming-Phoenix-2014T
  & 25 & 12.5 & 210$\times$260
  & de\_DE & Synthetic\ (same-signer concat) \\
Streaming-CSL-Daily
  & 30 & 15.0 & 512$\times$512
  & zh\_CN & Synthetic\ (same-signer concat) \\
\bottomrule
\end{tabular}
\end{table}

\begin{figure}[h] % Positioning: h = here
    \centering
    \includegraphics[width=0.9\textwidth]{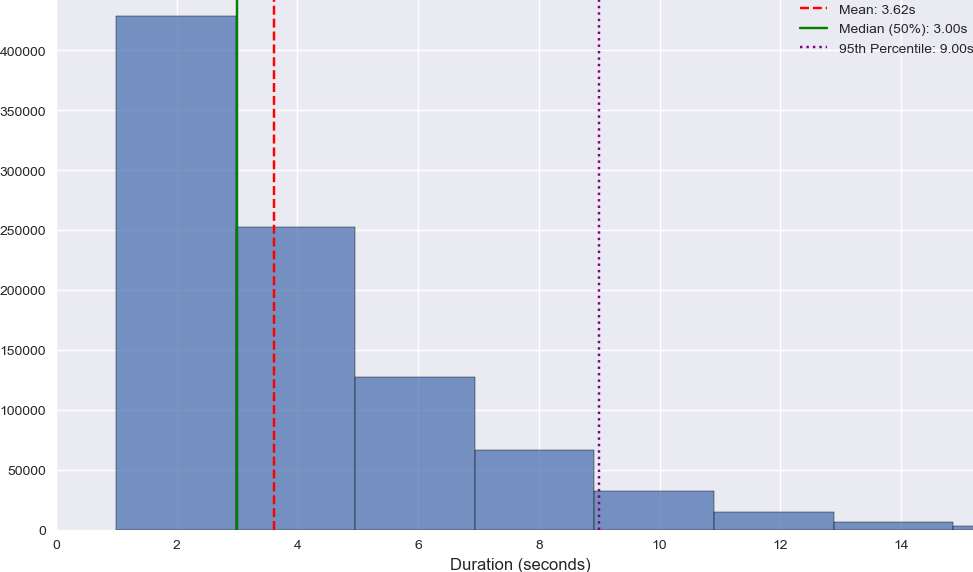}
    \caption{Distribution of subtitle durations in BOBSL. A mean of 3.6 seconds and a 95th percentile of 9 seconds, suggesting a 15-second window can typically contain 2–4 sentences.}
    \label{fig:bobsl}
\end{figure}

The 15-second figure matches the windowing protocol of~\citet{rao-etal-2025-sign}, which is the closest streaming-segmentation prior on the realistic \textbf{Streaming SLT} dataset (BOBSL), and is also informed by our subtitle-duration statistics shown in Figure~\ref{fig:bobsl}: mean BOBSL subtitle duration is 3.6 seconds with a 95th percentile of 9 seconds, so a 15-second window typically contains 2--4 sentences. Phoenix and CSL-Daily subtitles are shorter, so the same window holds proportionally more sentences. 

\subsection{Hungarian Matching} 
\label{sec:matching}

Following Deformable DETR~\citep{deformdetr,pdvc}, we use Hungarian matching~\citep{hungarian} to assign each ground-truth sentence to one of the $N$ candidate sentence queries before computing the final loss and before teacher forcing in the translation head.
For each candidate-target pair $(\hat{\mathbf{e}}_i, \mathbf{e}_j)$, the assignment cost is:
\begin{equation}
\mathcal{C}_{ij} = \lambda_{\text{cls}}^{m} \, \mathcal{C}_{\text{cls}}(\hat{y}_i, y_j) + \lambda_{L_1}^{m} \, \|\hat{\mathbf{b}}_i - \mathbf{b}_j\|_1 + \lambda_{\text{giou}}^{m} \,(1 - \text{GIoU}(\hat{\mathbf{b}}_i, \mathbf{b}_j)),
\end{equation}

where $\mathcal{C}_{\text{cls}}$ is the focal-style classification cost between predicted foreground confidence and the binary target (1 for sentences, 0 for background), and the localization cost combines $L_1$ distance and GIoU~\citep{giou}. The matching weights are $(\lambda_{\text{cls}}^{m}, \lambda_{L_1}^{m}, \lambda_{\text{giou}}^{m}) = (2, 0, 4)$.

\paragraph{Matching weights vs.~loss weights.} The matching weights above share the classification and GIoU weights of the loss in Equation~\ref{eq:joint_loss} of the main paper, $(\lambda_{\text{cls}}, \lambda_{\text{giou}}, \lambda_{\text{count}}, \lambda_{\text{trans}}) = (2, 4, 2, 2)$, and the $L_1$ term is inactive. A candidate that the matcher prefers is therefore also the candidate that the localization loss rewards. We follow the recipe of~\citep{pdvc} for both vectors. The Hungarian algorithm then solves $\sigma^* = \arg\min_\sigma \sum_i \mathcal{C}_{i\sigma(i)}$; matched candidates receive their corresponding ground-truth supervision, and unmatched candidates are trained as background under the focal loss only.

\subsection{Translation Configurations}

% All GT translations are tokenized using the pre-trained \textit{mBART-large-cc25} tokenizer. Each sequence is truncated or padded to a length of 40 tokens. This choice is sufficient to capture most BOBSL subtitles, which have a median length of approximately 8-10 words. 
% For the translation head, we also initialize the mBART decoder from the pre-trained \textit{mBART-large-cc25} model. 
% Following \cite{gfslt}, this module consists of three transformer decoder layers with 8 attention heads each, configured with a hidden dimension of 1024 and feed-forward dimension of 4096. 
% At inference, we generate translations autoregressively using greedy decoding, with each query producing a translation of up to 40 tokens. 

\paragraph{Per-language vocabulary trimming.} All ground-truth translations are tokenized using the pre-trained \textit{mBART-large-cc25} tokenizer~\citep{mbart}, which natively covers all three target languages in our benchmark suite (English, German, Chinese). To keep the captioner compact, we trim the tokenizer and decoder embedding matrix to the per-language vocabulary actually observed in each dataset, producing four lightweight per-benchmark decoders that share architecture and pre-trained weights but differ only in vocabulary size. Each sequence is truncated or padded to 40 tokens.

\paragraph{Decoder configuration.} Following~\cite{gfslt}, we use 3 transformer decoder layers with 8 attention heads, hidden dimension 1024, and feed-forward dimension 4096, initialized from the corresponding decoder layers of the trimmed \textit{mBART-large-cc25} checkpoint. At inference, we generate translations autoregressively using greedy decoding, with each query producing a translation of up to 40 tokens. 
% The full StreamSLST model contains 211.43M parameters -- two orders of magnitude smaller than the current pre-segmented SOTA on BOBSL~\citep{litfic}, which is built on a 8B-parameter Llama 3~\citep{llama}.

\subsection{Post-processing Strategy}

At inference time, the model produces a set of predictions per window. To select the most confident sentences, we compute a joint score of localization confidence and translation probability:
\begin{equation}
\text{score}_i = \alpha \cdot \frac{p_{\text{cap}}(c_i)}{|c_i|^T} + (1 - \alpha) \cdot p_i^{\text{cls}} \,,
\end{equation}
where $p_{\text{cap}}(c_i)$ is the product of token probabilities for generated translation $c_i$, $|c_i|$ is the translation length, $T = 2.0$ is a temperature parameter that normalizes for length bias, $p_i^{\text{cls}}$ is the foreground class probability or the detection confidence score, and $\alpha = 0.3$ balances the 2 components. This design ensures that sentences with both strong localization and coherent translations are prioritized, while penalizing overly short or verbose translations via the length normalization term. 

We rank all predictions by this score and select the top $K = 20$ sentences for evaluation, although in practice the predicted sentence count typically filters this to 1-6 sentences per window. Finally, we sort the selected sentences by their predicted start times to produce a temporally-ordered sequence of translations, suitable for display as streaming subtitles.

\subsection{Evaluation Metrics}
\label{app:metrics}

Following the dense video captioning protocol~\citep{pdvc}, all metrics are computed on the evaluation windows at each tIoU threshold in $\{0.3, 0.5, 0.7, 0.9\}$.

\paragraph{Localization.} Within a window, precision is the fraction of emitted predictions whose tIoU with a reference sentence reaches the threshold, and recall is the fraction of reference sentences that a prediction covers at this threshold. Both quantities count every prediction that the model emits, so a system that over-generates loses precision directly. We average precision and recall over windows and report $\mathrm{F1} = 2\bar{P}\bar{R} / (\bar{P} + \bar{R})$. The average F1 applies the same expression to $\bar{P}$ and $\bar{R}$ averaged over four thresholds, so it differs from the arithmetic mean of the four F1 values.

\paragraph{Translation.} Translation is scored on prediction--reference pairs. Each prediction is paired with every GT sentence whose tIoU reaches the threshold. Pairs are pooled within a stream, scored, and then averaged over streams and thresholds. Systems that share their intervals (e.g., \textbf{StreamSLST} and Cascade) are thus compared on identical pairs. Predictions from different windows are kept separate.

\subsection{Inference Cost and Latency}
\label{app:cost}

Table~\ref{tab:cost} reports the parameter count and the per-window cost of \textbf{StreamSLST} on each benchmark. The parameter counts differ only through the trimmed decoder vocabulary. GFLOPs grow with the vocabulary size, while the time per window stays between 222 and 229\,ms, because the decoder depth and the window length are fixed. One window is therefore processed in less than 2\% of its duration.

\begin{table}[h]
\centering
\caption{Per-window inference cost on one NVIDIA A100 at batch size 1 with greedy decoding, excluding pose extraction. GFLOPs assume a full 40-token decode for every query, and time is the median of 10 runs.}
\label{tab:cost}
\small
\begin{tabular}{l r c c c}
\toprule
\textbf{Benchmark} & \textbf{Vocab.} & \textbf{Params (M)} & \textbf{GFLOPs / window} & \textbf{ms / window} \\
\midrule
BOBSL & 26,286 & 211.4 & 216.3 & 226.5 \\
How2Sign & 10,708 & 179.5 & 183.8 & 229.2 \\
Streaming-Phoenix-2014T & 1,730 & 161.1 & 156.0 & 222.6 \\
Streaming-CSL-Daily & 5,792 & 169.5 & 171.7 & 223.8 \\
\bottomrule
\end{tabular}
\end{table}

\textbf{StreamSLST} emits a sentence once the window that fully contains it has arrived. Over all scored sentence--window pairs, the delay between the end of the sentence and the end of the window has a median of 6.0\,s and 5.9\,s and a 90th percentile of 10.8\,s and 11.8\,s on Streaming-Phoenix-2014T and How2Sign, respectively. Pose extraction runs frame by frame, so the temporal window is the only source of look-ahead.

\section{Representation Collapse Analysis}
\label{sec:collapse_analysis}

\begin{table}[h]
\centering
\caption{Variance across layers with/without VLP on Streaming-Phoenix-2014T and How2Sign.}
\label{tab:component_comparison}
\small
\begin{tabular}{lcccc}
\hline
\multirow{2}{*}{\textbf{Component}} 
  & \multicolumn{2}{c}{\textbf{Streaming-Phoenix-2014T}} 
  & \multicolumn{2}{c}{\textbf{How2Sign}} \\
\cmidrule(lr){2-3} \cmidrule(lr){4-5}
  & \textbf{Pre-train} & \textbf{No Pre-train} 
  & \textbf{Pre-train} & \textbf{No Pre-train} \\
\hline
\\
Input poses           & 0.026703 & 0.026703 & 0.017949 & 0.017949 \\
Backbone              & 3.242069 & 1.356921 & 3.949722 & 2.068815 \\
Encoder layer 1       & 0.948252 & 0.021566 & 0.938034 & 0.003247 \\
Encoder layer 2       & 0.722308 & 0.000389 & 0.225004 & 0.000217 \\
Decoder hidden states & 0.433345 & 0.000008 & 0.055886 & 0.000094 \\
Reference points      & 0.145557 & 0.000001 & 0.008492 & 0.000004 \\
\hline
\end{tabular}
\end{table}

% To further diagnose the gap in Table~\ref{tab:ablation1} (main paper), we measured the variance of hidden representations across layers in Table \ref{tab:component_comparison}. 
To further diagnose the gap reported in Table~\ref{tab:ablation1} of the main paper, we measured the layer-wise variance of hidden representations on the Streaming-Phoenix-2014T test set under both training strategies (Table~\ref{tab:component_comparison}). The collapse pattern is qualitatively identical when measured on How2Sign.
Without pretraining, variance collapses progressively through the network and critically at the reference points that determine sentence boundaries. Its near-zero variance (0.000001) indicates that all sentence queries produce nearly identical boundary proposals regardless of input content, a form of mode collapse where the model defaults to a single generic segmentation pattern. Contrastive pretraining prevents this collapse by grounding pose features in a multimodal embedding space before task-specific optimization, ensuring that encoder outputs retain sufficient diversity to support input-dependent localization and translation.

\section{Learned and Oracle Temporal Support}
\label{app:support}

This section examines the temporal support that \textbf{StreamSLST} learns and compares it with the annotated spans. In the oracle-boundary variant of Section~\ref{sec:ablation} (the first GT row of Table~\ref{tab:ablation2}), the decoder queries are built from annotated spans, boundary refinement is disabled, and the translation head is trained and evaluated on these spans. The variant reads the same 15-second multi-sentence windows as the full model, so two models differ in the source of the temporal support.

Table~\ref{tab:geometry} describes the learned spans that match an annotated sentence. For each sentence in a test window, we take the predicted span with the highest tIoU and keep the pair when this tIoU is positive. We then assign every frame of the learned span to its own sentence, to another sentence, or to no sentence. On Streaming-Phoenix-2014T, the learned spans stay close to the annotated spans: the median length ratio is 1.16, each boundary moves outward by about 0.4\,s on average, and the mean tIoU is 0.637. Most of a learned span lies on its sentence (81.6\%), and the added margins cover frames without a sentence (12.4\%) more than neighboring sentences (6.0\%). On How2Sign, the learned spans are wider (median ratio 2.25) and include more frames without a sentence (25.5\%) and more neighboring content (13.7\%).

\begin{table}[h]
\centering
\caption{Geometry and content of learned spans matched to annotated sentences (tIoU $>0$) on the test windows of the evaluation protocol. Offsets are learned minus annotated boundaries, so a negative start offset and a positive end offset widen the span.}
\label{tab:geometry}
\small
\begin{tabular}{l c c}
\toprule
\textbf{Statistic} & \textbf{Streaming-Phoenix-2014T} & \textbf{How2Sign} \\
\midrule
Annotated span, mean / median (s) & 3.63 / 3.05 & 3.85 / 3.00 \\
Learned span, mean / median (s) & 4.45 / 3.96 & 7.71 / 8.99 \\
Length ratio, mean / median & 1.40 / 1.16 & 3.21 / 2.25 \\
Start / end offset, mean (s) & $-0.42$ / $+0.40$ & $-1.64$ / $+2.22$ \\
tIoU, mean & 0.637 & 0.320 \\
Learned span longer than annotated (\%) & 65.7 & 80.3 \\
\midrule
Share on own / other / no sentence (\%) & 81.6 / 6.0 / 12.4 & 60.8 / 13.7 / 25.5 \\
\bottomrule
\end{tabular}
\end{table}

\paragraph{Interpretation with Table~\ref{tab:ablation2}} Joint optimization learns wider, translation-effective temporal support, and simple expansion or rescaling of annotated spans fails to reproduce its advantages. On Streaming-Phoenix-2014T, added margins consist mostly of frames without signed content, so the wider support adds little competing linguistic input. Exact boundaries are also an uncertain target on real video: on BOBSL, automatic subtitle timing achieves 46.98 average F1 against manual timing (Appendix~\ref{app:real_datasets}). A model that learns its own temporal support suits this setting, where the annotated boundaries are noisy in training and unavailable at test time. Our controls leave open whether learned proposals are semantically superior boundaries, and a unique cause of the advantage remains to be identified.

\section{Visual Front-End Control}
\label{app:frontend}

The gray rows in Tables~\ref{tab:loc_cap_bobsl_h2s} and~\ref{tab:loc_cap_phoenix_csl} list published systems that receive oracle boundaries on the original pre-segmented corpora and report corpus-level BLEU. They use a different task construction and a different aggregation, so we treat them as reference points. Table~\ref{tab:frontend} examines where our distance to them comes from.

\begin{table}[h]
\centering
\caption{Distance to pre-segmented systems. Panel A keeps the pre-segmented Phoenix-2014T task and corpus BLEU-4 and changes the visual front-end. Panel B reports Streaming-Phoenix-2014T under the matched-pair protocol of Appendix~\ref{app:metrics}. Each panel is internally comparable, and the two panels differ in task construction and BLEU aggregation.}
\label{tab:frontend}
\small
\begin{tabular}{l l l c}
\toprule
\textbf{System} & \textbf{Visual front-end} & \textbf{Input} & \textbf{BLEU-4} \\
\midrule
\multicolumn{4}{l}{\textit{Panel A: pre-segmented Phoenix-2014T, corpus BLEU-4 over all test clips}} \\
GFSLT-VLP~\citep{gfslt} (published) & RGB, 25\,fps, ResNet-18 & sentence clips & 21.44 \\
Sign2GPT~\citep{sign2gpt} (published) & RGB, 12.5\,fps, DINOv2 ViT-S/14 & sentence clips & 19.42 \\
GFSLT-VLP recipe, our front-end & pose, 12.5\,fps, CoSign~\citep{cosign} & sentence clips & 4.72 \\
\midrule
\multicolumn{4}{l}{\textit{Panel B: Streaming-Phoenix-2014T, matched-pair BLEU-4 averaged over four tIoU thresholds}} \\
\textbf{StreamSLST}, oracle support & pose, 12.5\,fps, CoSign~\citep{cosign} & 15\,s windows & 3.08 \\
\textbf{StreamSLST}, learned support & pose, 12.5\,fps, CoSign~\citep{cosign} & 15\,s windows & 8.16 \\
\bottomrule
\end{tabular}
\end{table}

In Panel A, we rerun the published GFSLT-VLP~\citep{gfslt} recipe on the pre-segmented Phoenix-2014T clips with our pose front-end in place of its RGB front-end. The result is 4.72 BLEU-4 against the published 21.44, which identifies the visual front-end as a major contributor to the gap under the pre-segmented protocol itself, before any boundary must be discovered. Our front-end differs from the published one in modality, frame rate, and encoder pretraining together, so the result characterizes the front-end as a whole. Uni-Sign~\citep{unisign} reaches competitive SLT results from pose input after language-specific pretraining on large paired corpora, such as 1{,}985 hours of Chinese sign language news video, whereas our encoder is trained on each target corpus alone. Panel B shows that oracle support yields 3.08 BLEU-4 under the streaming protocol. Perfect boundaries alone therefore leave the current translation stack well below the published pre-segmented systems, and Appendix~\ref{app:support} examines the difference between the two rows of this panel. The two panels follow different protocols, so we make no numerical decomposition across them.

\section{Case Studies}
\label{sec:case_study}

We provide qualitative examples on two of our four benchmarks, chosen for complementary purposes: Streaming-Phoenix-2014T (Sec.~\ref{sec:synth_case}) gives a controlled view of model behavior under exact GT boundaries, while BOBSL (Sec.~\ref{sec:bobsl_case}) illustrates a real-world failure mode of cascaded baselines.

\subsection{Synthetic Stream Case Studies (Streaming-Phoenix-2014T)}
\label{sec:synth_case}

We select Streaming-Phoenix-2014T for visual case studies because its sentence boundaries are exact by construction (Sec.~\ref{app:synthetic_datasets}), which lets us read off model behavior against ground truth without the auto-alignment noise that would confound a BOBSL example. The pose snapshots also serve a second purpose: they document what a synthesized stream actually \emph{looks like} after Hermite-bridge concatenation, which is otherwise only described algorithmically.

\begin{figure}[h]
    \centering
    \includegraphics[width=1.0\textwidth]{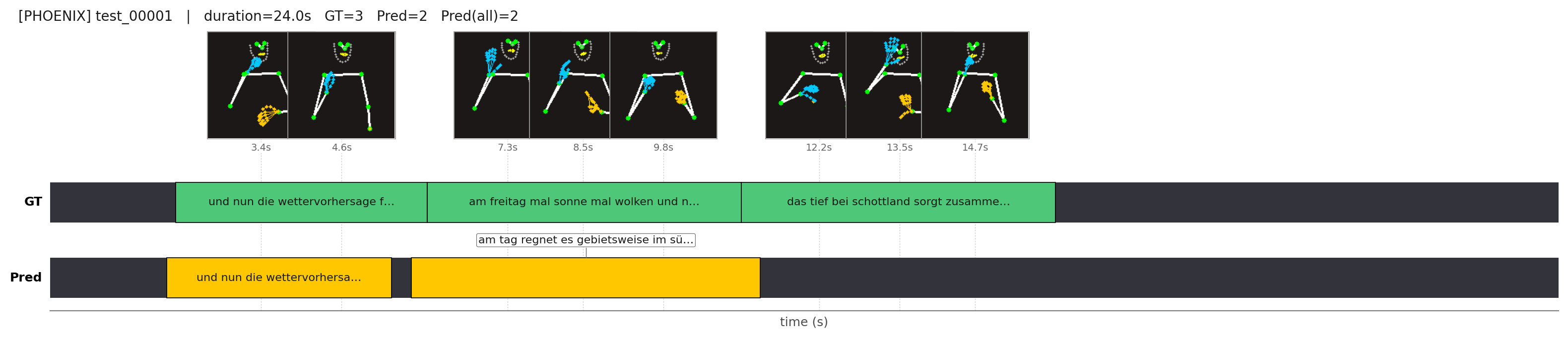}
    \caption{\textbf{Streaming-Phoenix-2014T} sample \texttt{test\_00001}. Pose snapshots are taken at three transition regions. Note that the hand pose flows continuously across each seam (e.g., 4.6\,s$\to$7.3\,s and 9.8\,s$\to$12.2\,s) rather than returning to a hands-down rest pose, validating our Hermite-bridge construction in Sec.~\ref{app:synthetic_datasets}. \textbf{StreamSLST} recovers the first two sentences with timing close to the ground truth (the first prediction begins at $\sim$2\,s, trimming the silent lead-in) but drops the third sentence at the stream tail. The first translation closely matches the reference. The second keeps the weather topic and changes its content, predicting rain (``\emph{am tag regnet es gebietsweise im sü...}'') where the reference describes sun and clouds on Friday (``\emph{am freitag mal sonne mal wolken}'').}
    \label{fig:case_phoenix_1}
\end{figure}

\begin{figure}[h]
    \centering
    \includegraphics[width=1.0\textwidth]{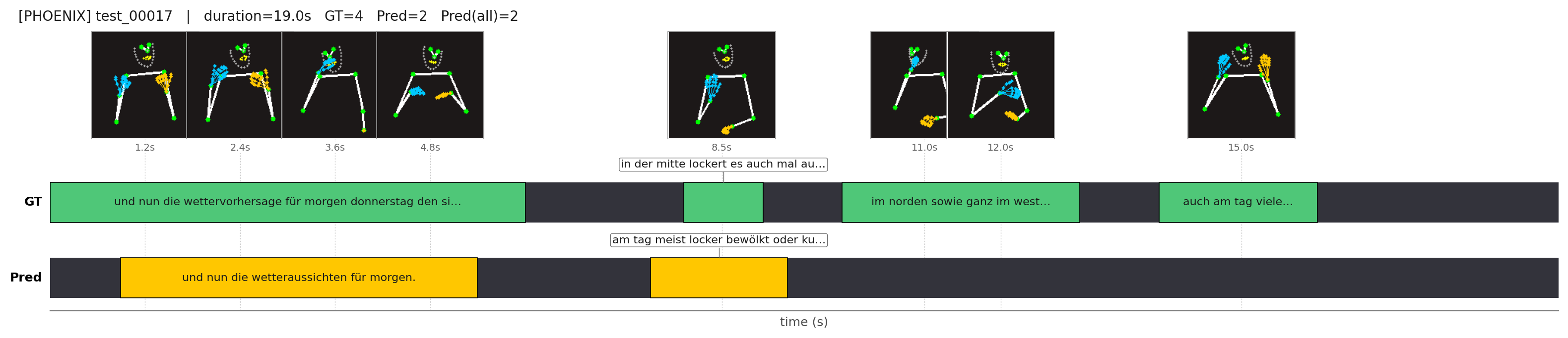}
    \caption{\textbf{Streaming-Phoenix-2014T} sample \texttt{test\_00017}. Two GT sentences have no predictions: the sentence at 10--13\,s (``\emph{im norden sowie ganz im west...}'') and the tail sentence at 15\,s. The first prediction keeps the meaning of the reference and omits the weekday, producing ``\emph{und nun die wetteraussichten für morgen}'' (and now the weather \emph{outlook} for tomorrow) for ``\emph{wettervorhersage für morgen donnerstag}'' (weather \emph{forecast} for tomorrow, Thursday). The second prediction covers the short sentence near 8.5\,s (``\emph{in der mitte lockert es auch mal au...}'') and changes its content (``\emph{am tag meist locker bewölkt oder ku...}'').}    
    \label{fig:case_phoenix_2}
\end{figure}

\paragraph{Three qualitative patterns} can be read off these examples and show the behavior that the main quantitative results summarize.

\textit{(i) The model emits fewer sentences than the stream contains.} Boundaries of detected sentences fall close to the reference, yet on \textbf{Streaming-Phoenix-2014T}, where boundaries are exact and the visual signal is uncorrupted, the model emits 2 of 3 sentences in Figure~\ref{fig:case_phoenix_1} and 2 of 4 sentences in Figure~\ref{fig:case_phoenix_2}.

\textit{(ii) The missed sentences are the last sentences of the stream.} In both streams, the last sentence crosses the end of the 15-second evaluation window, so the model reads only part of it. In Figure~\ref{fig:case_phoenix_2}, the model also misses the sentence before it, which lies inside the window, so the \textit{Sentence Counter} undercounts by one there. Padding at the stream end and a content-conditioned foreground threshold are two direct extensions.

\textit{(iii) The translation head paraphrases rather than transcribes.} Some outputs match the reference closely, some keep its meaning while omitting a detail such as the weekday, and some keep only its topic, and BLEURT scores the last two kinds more leniently than BLEU-4 does. For example, substituting ``\emph{wetteraussichten}'' (outlook) for the reference ``\emph{wettervorhersage}'' (forecast) in Figure~\ref{fig:case_phoenix_2} is the precise kind of edit that pulls BLEU-4 down (zero 4-gram overlap is preserved through such a swap) while leaving BLEURT close to the reference (the two sentences are valid paraphrases of one another). The second prediction in Figure~\ref{fig:case_phoenix_1} is a content error instead: it is a fluent weather statement whose content differs from the reference, and it remains topically close to it. A high BLEURT beside a low BLEU-4 can therefore reflect topical similarity as well as paraphrase. The same tendency at the corpus level is what creates the wide BLEU-4 spread and the narrow BLEURT spread that we report across the four datasets in Sec.~\ref{sec:results}. The mBART decoder produces fluent target-language text that captures the topic of the signed content rather than the exact subtitle string. A paraphrase keeps the meaning for a reader, whereas exact-match metrics penalize it, which is why Section~\ref{sec:results} argues that BLEURT and METEOR should be the primary cross-language indicators on \textbf{Streaming SLT}.

\subsection{Cascaded vs.~Joint Failure Mode}
\label{sec:bobsl_case}

\begin{table}[h]
    \centering
    \caption{\textbf{Qualitative comparison on a continuous stream.}
    Cascade baseline fails due to error propagation from the segmentation module (red), resulting in fragmented and meaningless translations.
    In contrast, our \textbf{StreamSLST} (blue) learns robust temporal boundaries that preserve semantic integrity of signing sentences, generating coherent translations even under complex streaming conditions.}
    \label{tab:case_study}
    \renewcommand{\arraystretch}{1.3}
    \resizebox{\textwidth}{!}{
    \begin{tabular}{c|l|l}
        \toprule
        \textbf{Model} & \textbf{Timeline (s)} & \textbf{Translation Output} \\
        \midrule
        \rowcolor{gray!10} \multicolumn{3}{l}{\textit{Input Stream Interval: 00:00:47 $\sim$ 00:01:00} [Vid: 5242317681679687839]} \\
        \midrule
        \multirow{2}{*}{\textbf{\shortstack{Ground\\Truth}}}
        & 47.7 -- 52.9 & A rainy August has put things badly behind, so now, there's an awful lot at stake. \\
        & 52.9 -- 58.8 & A good harvest means a good year for farmers, a bad one can spell disaster. \\
        \midrule
        \multirow{3}{*}{\textbf{\shortstack{Cascade\\Baseline}}} 
        & 48.5 -- 50.1 \textcolor{red}{(Clipped)} & Rainy August behind. \textcolor{red}{\xmark (Missed Context)} \\
        & 51.2 -- 52.0 \textcolor{red}{(Frag.)} & Now lot. \textcolor{red}{\xmark (Broken)} \\
        & 54.0 -- 57.5 \textcolor{red}{(Late)} & Year farmers, bad disaster. \textcolor{red}{\xmark (Missed Subject)} \\
        \midrule
        \multirow{2}{*}{\textbf{\shortstack{StreamSLST\\(Ours)}}} 
        & 47.4 -- 53.1 \textcolor{blue}{(Learned)} & Because of a rainy August, things are behind and a lot is at stake. \textcolor{green}{\cmark} \\
        & 53.2 -- 59.0 \textcolor{blue}{(Learned)} & A good harvest implies a good year, but a bad one means disaster. \textcolor{green}{\cmark} \\
        \bottomrule
    \end{tabular}
    }
\end{table}

To visually examine the impact of joint optimization, we provide a case study in Table \ref{tab:case_study} on a continuous stream containing two causally linked sentences. Cascade baseline (middle row) demonstrates the fragility of disjoint pipelines. The segmentation module fails to handle the continuous transition, erroneously fragmenting the first sentence into disconnected clips (``Clipped'' and ``Frag'') and triggering the second sentence too late (``Late''). Consequently, the translation model receives incomplete visual contexts, leading to severe semantic loss (e.g., missing the subject ``harvest'' and the causal link ``so now''). This confirms that segmentation errors in streaming scenarios cause irreversible error propagation.

In contrast, \textbf{StreamSLST} (bottom row) successfully recovers the full semantic content. Notably, the model predicts learned boundaries (e.g., 47.4--53.1s) that slightly differ from the GT but effectively encapsulate the complete linguistic unit. By jointly optimizing segmentation and translation, \textbf{StreamSLST} maintains the logical flow (``Because of...'', ``implies...''), demonstrating that the model prioritizes semantic completeness over rigid temporal alignment.

\section{Limitations and Future Work}
\label{sec:limitations_future}

While \textbf{StreamSLST} establishes a strong baseline for \textbf{Streaming SLT}, it represents an initial step towards solving this complex problem. We identify several limitations and outline promising directions for future research.

\subsection{Limitations}

\textbf{Translation quality lags pre-segmented systems.} 
BLEU-4 stays well below that of systems that consume oracle boundaries on pre-segmented corpora, and our front-end control identifies the leading cause (Appendix~\ref{app:frontend}). Our encoder reads 2D pose at half the source frame rate and learns from the target corpus alone, whereas stronger pose-based systems depend on large-scale paired pre-training that has no counterpart for BSL or DGS. A second cause is decoder capacity, since our trimmed three-layer mBART decoder is smaller than the LLM decoders used by recent pre-segmented systems (\citep{litfic}, \citep{sign2gpt}, \citep{mmslt}). A third cause is the streaming setting itself, where an occasional boundary miss propagates into a missing or fragmented translation. We regard a large-scale pretrained pose encoder as the first direction for improvement (Section~\ref{sec:future_directions}).

\textbf{Window-level inference.}
Inference in \textbf{StreamSLST} treats each 15-second window on its own and carries no state across windows. A sentence becomes available once the window that contains it ends, so the emission delay depends on the window length and the stride. A stateful buffer that carries evidence across windows is therefore required to merge predictions and shorten the delay.

\textbf{Sensitivity to upstream pose quality.} 
Because the only visual signal reaching the encoder is the 2D pose stream, errors in the pose extractor propagate end-to-end. On BOBSL we observe degraded output in segments where DWPose~\citep{dwpose} fails to recover hand keypoints (low-light scenes, occlusions, off-camera signers); on How2Sign the OpenPose-derived keypoints are noisier than DWPose, which we believe contributes to the smaller pretraining benefit on H2S relative to Phoenix (Table~\ref{tab:ablation1}). The framework has no mechanism to ``recover'' from systematically corrupted pose input.

\textbf{Synthetic-stream realism gap.}
The two synthetic benchmarks concatenate same-signer sentence clips with interpolated bridges (Appendix~\ref{app:synthetic_datasets}). This construction preserves the signing within each clip and gives exact boundaries, while its transitions and pauses follow the construction rules rather than natural discourse. Claims about discourse-level phenomena therefore rest on the real corpora. 

\textbf{Scope of the real data.}
The real corpora contain interpreted broadcast signing (BOBSL) and scripted instructional signing (How2Sign). They include no dialogue, turn-taking, or spontaneous conversation between Deaf signers. Conversational corpora such as the BSL Corpus~\citep{bslcorpus} record one video view per signer, which matches the single-signer input of \textbf{StreamLST}, and they are the natural next benchmark for our \textbf{Streaming SLT} task. 

\subsection{Future Directions}
\label{sec:future_directions}

\textbf{Large-Scale Pose Pretraining.}
Our front-end control in Appendix~\ref{app:frontend} attributes much of the gap to the visual front end, so we take large-scale pretraining of the pose encoder as the first direction for improvement. Pose-based systems reach competitive gloss-free translation when a large paired pretraining corpus is available, such as the 1{,}985 hours of CSL-News in Uni-Sign~\citep{unisign}, whereas our encoder is trained on each target corpus alone. Pretraining the pose backbone at this scale is expected to improve translation quality in both the pre-segmented and streaming settings.

\textbf{Multi-Modal Fusion.} 
To address the modality limitation, future work should explore fusing pose features with lightweight RGB representations. Integrating visual encoders (e.g., adapters used in Uni-Sign or pre-trained vision backbones) could recover missing details such as facial expressions and object interactions, thereby enhancing translation fidelity without significantly compromising inference speed.

\textbf{LLM-Integrated Decoding.}
Currently, \textbf{StreamSLST} employs a standard mBART decoder. Given the recent success of Large Language Models (LLMs)~\citep{guo2026beyond,guo2026ssmnbench,zhu2026video} in gloss-free translation, replacing the translation head with a visually-conditioned LLM (e.g., Llama 3 or Qwen) represents a significant opportunity. LLMs could leverage their powerful world knowledge and reasoning capabilities to generate more coherent and contextually accurate translations from the noisy visual signals.

\textbf{Optimization for Real-Time Inference.}
Our ultimate goal is deployment in real-world streaming scenarios. Future efforts will focus on system-level optimizations, such as converting the sliding window mechanism into a stateful streaming buffer and employing quantization techniques. This would minimize latency and enable the deployment of \textbf{StreamSLST} on edge devices for accessible daily communication.

%%%%%%%%%%%%%%%%%%%%%%%%%%%%%%%%%%%%%%%%%%%%%%%%%%%%%%%%%%%%

% \newpage
% \input{checklist.tex}

\end{document}